%% file: paper.tex
\pdfoutput=1
\documentclass[11pt]{article}
\PassOptionsToPackage{table}{xcolor}

\usepackage{iclr2027_conference}
\iclrfinalcopy

\usepackage{times}
\input{math_commands.tex}

\usepackage{latexsym}
\usepackage{tikz}
\usetikzlibrary{positioning,fit,calc,backgrounds}
\usepackage[T1]{fontenc}
\usepackage[utf8]{inputenc}
\usepackage{microtype}
\IfFileExists{inconsolata.sty}{\usepackage{inconsolata}}{}
\usepackage{graphicx}
\usepackage{amsmath}
\usepackage{caption}
\usepackage{booktabs}
\usepackage{placeins}
\usepackage{xcolor}
\usepackage{enumitem}
\usepackage{float}
\usepackage{listings}
\definecolor{mplgreen}{HTML}{2CA02C}
\definecolor{mplred}{HTML}{D62728}
\definecolor{mplblue}{HTML}{1F77B4}
\definecolor{bandcol}{HTML}{E8E8E8}
\definecolor{meanband}{HTML}{E8E8E8}
\definecolor{qwenband}{HTML}{EEF3F8}
\definecolor{gemmaband}{HTML}{FBF6E8}
\definecolor{ministralband}{HTML}{EFF7F0}

\newcommand{\ci}[2]{\begin{tabular}[c]{@{}c@{}}#1\\[-2pt]{\tiny #2}\end{tabular}}

\title{Instruction Retrieval at Inference Time for Small Language Models}

\author{Kenan Alkiek, David Jurgens \& V.G.Vinod Vydiswaran \\
School of Information \\
University of Michigan \\
Ann Arbor, MI 48109, USA \\
\texttt{\{kalkiek,jurgens,vgvinodv\}@umich.edu} \\
}

\begin{document}

\maketitle
\lhead{Preprint. Under review.}

\begin{abstract}
The facts a language model stores are tied to its parameter count, so small models that fit on edge devices fail on expert problems, which need specialized knowledge and follow multi-step procedures. Fine-tuning for a specific domain or task writes the knowledge into the parameters but must be repeated for every model and domain, and a retrieved passage leaves the model to find the relevant fact and apply it on its own. We introduce instruction retrieval, which distills a teacher model's expertise into a corpus of instructions tailored so that a small model can follow. For each cluster of a domain's problems, the teacher writes one instruction with the background knowledge the cluster depends on, a procedure for that kind of problem, and the common mistakes made on it. This reusable corpus needs only to be built once per domain and requires no run-time teacher access.
At inference, a frozen small model retrieves the instructions nearest its question and follows them, with no fine-tuning. Across medicine, law, and mathematics benchmarks, we demonstrate the corpus improves over zero-shot on every task and over few-shot prompting, self-consistency, and other retrieved text on medicine and law. On MedQA the corpus raises mean accuracy by 10.6 points, where retrieved textbook passages raise it by 3.9 and few-shot examples from the same teacher lower it. An error analysis shows that only the background knowledge fixes the questions a small model always gets wrong, and that the procedure and common mistakes fix only the questions where it wavers between options. Our results show that automatically retrieved inference-time procedural guidance and domain knowledge can yield substantial gains for small models.
\end{abstract}

\section{Introduction}

Large language models reason at an expert level across professional domains, from clinical diagnosis \citep{kwon2024largelanguagemodelsclinical, singhal2025expert} to bar examinations \citep{katz2024bar} to competition mathematics \citep{hendrycks2021measuringmathematicalproblemsolving, deepseekai2025deepseekr1incentivizingreasoningcapability}.  A 175B-parameter model occupies over 300GB of GPU memory in half precision and runs only across several datacenter GPUs \citep{pmlr-v202-xiao23c,frantar2023gptqaccurateposttrainingquantization}, whereas a small language model (SLM) of a few billion parameters runs on a laptop or a phone at a fraction of the cost \citep{pham2024slimlmefficientsmalllanguage,belcak2025smalllanguagemodelsfuture}. Small models are also improving quickly, and across open releases the number of parameters needed to reach a fixed benchmark score halves about every 3.5 months \citep{xiao2025densing}. For example, Gemma 3's 4B instruction-tuned model is competitive with the 27B model of the generation before it \citep{gemma3_2025}, and Gemma 4's E2B model reports results near Gemma 3's 27B \citep{gemmateam2026gemma4}.

What has not changed is how much an SLM can memorize. The facts a model stores are tied to its parameter count \citep{roberts-etal-2020-much,lu-etal-2024-scaling}, so a small model holds a fraction of the world knowledge of a large one by design, regardless of what it is trained on. The lack of world knowledge particularly affects problems that require specialized knowledge about a domain and the multi-step procedures for applying it to the domain's problems. A standard remedy is to fine-tune an SLM on the target domain, which instills the missing knowledge into the parameters. However, while fine-tuning raises accuracy on that domain, it trades general ability for a target skill \citep{fu2023specializingsmallerlanguagemodels, kotha2024iclr-understanding} and must be repeated for every new model or domain. The other remedy stores the knowledge outside the model. Retrieval-augmented generation (RAG) retrieves a handful of passages from an external corpus and places them in the prompt \citep{lewis2021retrievalaugmentedgenerationknowledgeintensivenlp}, with no training of the model \citep{ram-etal-2023-context}. However, a retrieved passage is general text, so the model must find the relevant fact inside it and apply it on its own; further, these passages rarely tell models what to do with the retrieved facts and thus do not fill in procedural knowledge. Small models do this less well than large ones, and answer wrong even when a retrieved passage contains what they need \citep{joren2025iclr-sufficient}, gaining more from retrieved passages once they are fine-tuned to read them \citep{zhang2024raft,asai2024iclr-selfrag}. A small model therefore needs the missing knowledge in a form it can follow and relevant procedures for applying it.

Here, we introduce \emph{instruction retrieval}, distillation from a teacher model into a corpus rather than into weights (Figure~\ref{fig:corpus}). We cluster a domain's problems, and for each cluster a teacher writes one instruction with the background knowledge, a procedure, and the common mistakes for that type of problem. At inference, a frozen SLM retrieves the knowledge-grounded instructions most similar to its question and follows them.
In experiments with medicine, law, and mathematics benchmarks, using the corpus raises accuracy on each. On MedQA, it raises the mean accuracy of five small models by 10.6 points over zero-shot, where retrieved textbook passages raise it by 3.9 and few-shot examples from the same teacher lower it; the gain is largest on medicine, and largest there for the smallest models.

Our contributions are threefold. First, we present instruction retrieval, a method for distilling a teacher's expertise into a corpus of instructions written for small models to retrieve and follow (Section~\ref{sec:corpus}). Second, across three domains and five small models from three families, the corpus improves over zero-shot on every task and over every baseline that generates one answer (Section~\ref{sec:results}). Third, an error analysis shows that only the background knowledge reaches the questions a small model gets wrong on every attempt, while the procedure and common mistakes fix only the questions where it wavers between options (Section~\ref{sec:analysis}). All code and data are in the supplementary material and will be released on publication.

\begin{figure}[t]
\centering
\input{figures/figure1}
\caption{One board-exam question with the text a small model reads under passage retrieval and under instruction retrieval. The passages are textbook pages on ulcerative colitis; the fact the question needs (highlighted in both panels) is one sentence among many. The instruction states that fact as background knowledge, orders the steps in the procedure (a plain x-ray first, CT only if the diagnosis stays unclear), and names the common mistakes, ordering CT or colonoscopy first. On MedQA, instructions raise mean accuracy by 10.6 points and passages by 3.9 (Table~\ref{tab:main}). The full instruction is in Appendix~\ref{app:instruction}.}
\label{fig:corpus}
\end{figure}

\section{Related Work} \label{sec:related}

A small model can be given expertise it does not hold by distilling a teacher into its weights, by retrieving text into its prompt, or by writing guidance from its own errors.

\paragraph{Distillation} Reasoning distillation, the standard route to a small model that solves expert problems, fine-tunes the student on traces a large teacher generates \citep{ho2023largelanguagemodelsreasoning,li2022explanationslargelanguagemodels,magister2023teachingsmalllanguagemodels,hsieh2023distillingstepbystepoutperforminglarger}, with gains reported on mathematical benchmarks such as GSM8K \citep{cobbe2021trainingverifierssolvemath} and MATH \citep{hendrycks2021measuringmathematicalproblemsolving}. Later variants distill structured thought templates. SuperCorrect fine-tunes 7--8B students on teacher-written templates \citep{yang2025supercorrect}, and ReasonFlux trains a navigator model to plan over a teacher-written template library \citep{yang2025reasonflux}. KARD adds retrieval to distillation for knowledge-intensive tasks, including MedQA \citep{kang2023knowledgeaugmentedreasoningdistillationsmall}, as does Probe-then-Retrieve-and-Reason \citep{zhao-etal-2024-probe}. However, each of these methods fine-tunes every student it serves, and fine-tuning specializes the student, spending its limited capacity on one target ability at a measured cost in generality \citep{fu2023specializingsmallerlanguagemodels} and repeating the training for every new student and domain. Instruction retrieval keeps the student frozen and general, and stores the teacher's expertise as text that is written once and followed by any student.

\paragraph{Retrieval augmentation} Dense retrieval gives a model facts it lacks by placing passages in the prompt \citep{karpukhin-etal-2020-dense,lewis2021retrievalaugmentedgenerationknowledgeintensivenlp}, including passages from medical textbooks \citep{wang2025augmentingblackboxllmsmedical}, and the model still has to find the fact in the passage and reason from it. Context written for the question by a large model can match or outperform retrieved passages on open-domain and commonsense questions \citep{liu2022generatedknowledge,yu2023generateretrievelargelanguage}, but the text is written per question by the model that answers, and discarded. Retrieval can also supply reasoning. Buffer of Thoughts distills thought templates from the solver's own solution traces \citep{yang2024bufferofthoughts}, Self-Discover has the solver compose a task-level reasoning structure from generic seed modules \citep{zhou2024selfdiscover}, and Hypotheses-to-Theories induces a rule library from training examples and prepends it whole at inference \citep{zhu2024learnrules}. ToTAL places the entire template library in the prompt \citep{jeong2025thoughttemplates}, Retrieval-of-Thought reuses template steps to shorten the traces of thinking-mode models \citep{ahmed2025retrievalofthought}, and retrieval over another model's thinking traces improves frontier solvers \citep{arabzadeh2026ragthinking}. All six make a model that already solves the problem more efficient, because the guidance shortens or reuses reasoning the solver could produce itself, or is written by the solver. Our students fail these problems outright and cannot write the guidance, so a teacher writes it for them, and what it writes is the knowledge and the mistakes the student lacks rather than a shortened trace of reasoning it already has.

\paragraph{Error-driven corpora} Three recent methods write guidance from a model's own mistakes. RICP has a teacher write a reason and insights for each training error of the student, clusters the errors into task-level principles shared by every question, and retrieves per question the insights of the most similar errors \citep{sun-etal-2024-retrieved}. Mistake Notebook Learning has the agent itself cluster its failures and distill each cluster into a note \citep{su2025mistake}. Contrastive Reflection Memory stores pairs of correct and incorrect reasoning and retrieves them at inference \citep{li2026trainingfree}. RICP shares the teacher-writes, student-retrieves design and is the baseline in Section~\ref{sec:results}. However, a corpus written from one model's errors covers only the questions that model got wrong and can't be reused. The Instruction Corpus covers every cluster of a domain's problems, and the students' errors are used only to write the common mistakes inside each instruction.

\section{Instruction Corpus} \label{sec:corpus}

We build an \emph{Instruction Corpus} for each domain, a collection of teacher-written instructions that a frozen student retrieves and follows at inference (Figure~\ref{fig:corpus}). Each instruction is written for one cluster of related problems and has three parts. The background knowledge states the facts the cluster's problems depend on. The procedure gives the steps for applying those facts to that kind of problem, for a model that holds the facts and does not know how to apply them. The common mistakes name what separates the option students chose from the correct one, as prior work does when it writes a corpus from a model's own errors \citep{sun-etal-2024-retrieved,su2025mistake,li2026trainingfree}.

For the question in Figure~\ref{fig:corpus}, the background knowledge states the facts the answer needs (i.e., that toxic megacolon is confirmed on a plain abdominal x-ray and that colonoscopy and contrast enemas are contraindicated). The procedure orders the steps, an x-ray first and CT only if the diagnosis stays unclear. The common mistakes name CT, the test the students chose zero-shot, as the test not to order first. 
Formally, we define the Instruction Corpus of a domain $d$ as $\mathcal{C}_d = \{I_1, \ldots, I_N\}$, one instruction per cluster, and each instruction is a triple $I_j = (K_j, P_j, M_j)$, where $K_j$ is the set of background-knowledge statements, $P_j$ an ordered list of procedure steps that apply that knowledge to problems of the cluster's type, and $M_j$ a set of common mistakes to avoid. Section~\ref{sec:interventions} measures which part fixes which kind of wrong answer.

\paragraph{Corpus construction} A corpus is built from a set of problems of the domain, which can come from any source, such as problems a language model writes from a body of domain text, problems mined from a search log, or problems posted to a public forum. Here, the problems are each benchmark's training questions. We build the corpus once per domain in two steps. First, we embed the training questions with OpenAI's \texttt{text-embedding-3-large} and group them by agglomerative clustering with average linkage and cosine distance. The cluster size sets how broad a group each instruction covers, where a coarse cut yields broad categories and a fine cut yields specific problem types. We select the cluster size on the development split, three questions per cluster, and the sweep is in Appendix~\ref{app:selection}. Second, for each cluster $j$ the teacher, GPT-5, writes one instruction $I_j$. $K_j$ and $P_j$ are written from the cluster's training questions and their answers, and $M_j$ from the wrong answers the students gave on those questions, pooled over the four Qwen and Gemma students, so one corpus per domain serves every student. $N$ is 3{,}393 on MedQA, 333 on MMLU Law, and 9{,}761 on MathQA, and the mean instruction is 280 words; the teacher and student prompts are in Appendix~\ref{prompt-templates}.

\paragraph{Retrieval} At inference the student reads the instructions written for the training questions most like its question $q$. We embed $q$ with the same encoder, rank each $I_j$ by the mean cosine similarity between the embedding of $q$ and the embeddings of the three training questions $I_j$ was written for, and place the five nearest instructions in the prompt. The teacher is not called at inference. The student reads the retrieved instructions and answers. The encoder, the number retrieved, and the position in the prompt are the same as for the passage-retrieval baseline (Section~\ref{sec:setup}); the number retrieved is chosen on the development split (Appendix~\ref{app:selection}).

\paragraph{Corpus validation} The Instruction Corpus is built from the training split alone, as a fine-tuned model is, and the teacher sees no test question. The teacher is a frontier model that may have seen these benchmarks in pretraining, as any teacher in distillation may, and it could restate a test question from memory inside an instruction. We therefore check every $I_j \in \mathcal{C}_d$ against the test split with the 13-gram overlap probe used for contamination \citep{brown2020languagemodelsfewshotlearners}, and repeat it at 8 grams, the length PaLM and Llama 3 use \citep{chowdhery2023palm,grattafiori2024llama3herd}. No test question on any of the three benchmarks shares a 13-gram with any instruction. At 8 grams, 0.2\% of MedQA, 0.4\% of MMLU Law, and 0.4\% of MathQA test questions share a phrase with an instruction. We read every match, and each is a stock phrase of the domain such as ``holosystolic murmur at the left lower sternal border'', not a restated question (Appendix~\ref{app:decontam}).

A separate question is whether an instruction gives the student the answer itself. We give each student the five retrieved instructions and the answer options with the question withheld, and ask the student to choose an option. We compare the rate at which the students choose the keyed option against the rate when they see only the options, and on MedQA against the rate when they see the five retrieved textbook passages instead, which are not written to any question. On MedQA the students choose the keyed option 38.4\% of the time from the instructions, 39.8\% from the passages, and 22.7\% from the options alone; with the question, 74.9\%. On MMLU Law the rate is 39.8\% from the instructions against 34.2\% from the options alone, and on MathQA 10.3\% against 9.5\% (Appendix~\ref{app:decontam}). An instruction therefore carries what its cluster of questions has in common, about as much as a textbook passage does, and not the answer to any one question.

\section{Experimental Setup} \label{sec:setup}

We evaluate the corpus on medicine, law, and mathematics with five small students from three model families, against the standard inference-time baselines. Every run uses the same student prompt and output format, so the conditions differ only in the text placed in the prompt or in the number of answers generated.

\paragraph{Tasks} We use three benchmarks that ask for different kinds of expertise. MedQA \citep{jin2020diseasedoespatienthave} contains multiple-choice questions from medical board exams; success requires recall of biomedical facts and diagnostic reasoning. MMLU Professional Law \citep{hendrycks2021measuringmassivemultitasklanguage}, hereafter MMLU Law, contains professional-law questions, testing recall of rules and precedent alongside case-based reasoning. MathQA \citep{amini2019mathqainterpretablemathword} contains math word problems, which require translating text into a sequence of arithmetic operations. MedQA uses its published splits of 10{,}178 / 1{,}272 / 1{,}273 questions for training, development, and test. MMLU Law has no training split and a 172-question test split, too small to separate conditions a few points apart, so we re-split its 1{,}709 questions into 999 / 170 / 540. For MathQA we use the published training and test splits and draw a 1{,}000-question development split from the published development split, after removing training questions that duplicate a development or test question, leaving 29{,}283 / 1{,}000 / 2{,}975.

\paragraph{Students and teacher} We evaluate five small models from three families: Qwen 3.5 2B and 4B \citep{qwenteam2026qwen35}, Gemma 4 E2B and E4B \citep{gemmateam2026gemma4}, and Ministral 3 3B \citep{mistral2026ministral3}. We run every student with its thinking mode off, because the output format already asks for step-by-step reasoning; the Qwen and Gemma students have a thinking mode, which we return to in Section~\ref{sec:interventions}. GPT-5 (\texttt{gpt-5-2025-08-07}) \citep{openai2025gpt5} writes every instruction, and we also report its zero-shot accuracy on the test questions as a frontier reference. Checkpoints and inference settings are in Appendix~\ref{appendix-inference}.

\paragraph{Baselines} We compare against five baselines and one control. Zero-shot prompting presents the question in the same output format with no added context. Because the format asks for step-by-step reasoning before the answer, this baseline already includes chain-of-thought. Few-shot prompting adds five worked examples per benchmark, written by the same teacher, the same five for every question \citep{brown2020languagemodelsfewshotlearners,wei2023chainofthoughtpromptingelicitsreasoning}. Self-consistency samples three responses and takes the majority vote \citep{wang2023selfconsistencyimproveschainthought}. Retrieved In-Context Principles (RICP) \citep{sun-etal-2024-retrieved} retrieves principles that the teacher wrote from the students' mistakes on the training split; as in our method, the teacher writes and the student retrieves. Passage retrieval, on MedQA only, adds the five passages nearest the question from the benchmark's medical textbooks \citep{wang2025augmentingblackboxllmsmedical}, retrieved with the same embedder. The other two benchmarks provide no comparable text corpus. The control, random instructions, places five instructions drawn at random from $\mathcal{C}_d$, so that the gain from retrieving the nearest instructions is separated from the gain from any instruction text.

\paragraph{Decoding and metrics} We use greedy decoding for every condition except self-consistency, which samples three responses at temperature 0.7 and top-p 0.95. We report accuracy on the test split; paired bootstrap confidence intervals for every condition's difference from zero-shot are in Appendix~\ref{app:full-results}. A format failure, an answer that cannot be extracted from the output, is scored wrong. Zero-shot, the rate is at most 1.5\% for every student and task except Qwen 3.5 2B and Ministral 3 3B on MathQA, at 8.6\% and 5.8\%, where the student keeps reasoning and never states an answer (every condition in Appendix~\ref{app:main-full}).

\begin{figure}[t]
\centering
\includegraphics[width=\textwidth]{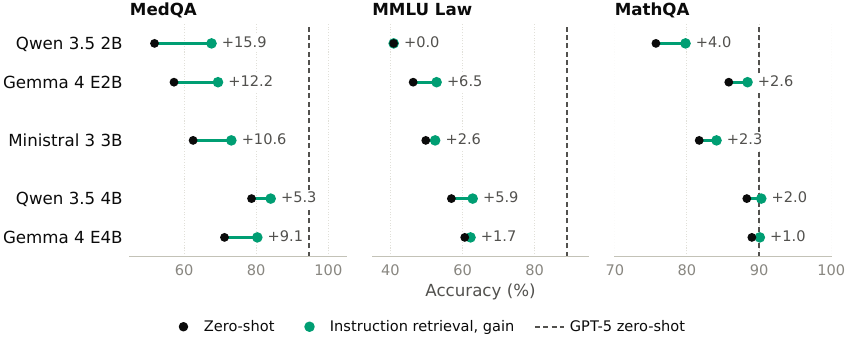}
\caption{Zero-shot and instruction-retrieval accuracy for each student on the three test splits. Each row is one student; the black dot is its zero-shot accuracy, the green dot its accuracy with the corpus, the label the gain in points, and the dashed line GPT-5 zero-shot, the frontier reference. No student loses accuracy on any task, and every student gains on MedQA and MathQA, most on MedQA, where the two 2B students gain 16 and 12 points.}
\label{fig:main}
\end{figure}

\section{Results} \label{sec:results}

We evaluate the corpus on the test splits against the baselines of Section~\ref{sec:setup}. Table~\ref{tab:main} reports the results for every student and task, and Figure~\ref{fig:main} shows each student's gain from zero-shot to the corpus against the frontier reference. GPT-5 zero-shot, the reference, scores 94.7\% on MedQA, 88.9\% on MMLU Law, and 90.0\% on MathQA, so the students' zero-shot means stand 30.4, 38.0, and 5.9 points below it.

\paragraph{The corpus raises accuracy on every task, most on MedQA, where the smallest students gain most.} A student that reads the corpus answers questions it could not answer alone. Over the five students, the corpus raises the MedQA mean from 64.3 to 74.9\%, MMLU Law from 50.9 to 54.2\%, and MathQA from 84.1 to 86.5\%. The gain is outside its 95\% bootstrap interval for every student on MedQA and MathQA, and on MMLU Law for Qwen 3.5 4B and Gemma 4 E2B (Appendix~\ref{app:full-results}). The smaller student in each family gains most on MedQA, 15.9 points for Qwen 3.5 2B and 12.2 for Gemma 4 E2B against 5.3 and 9.1 for the 4B sizes, because the larger student in each family gets fewer questions wrong on every attempt, 8\% against 17\% for Qwen 3.5 and 15\% against 25\% for Gemma 4 on MedQA (Section~\ref{sec:errors}). On MMLU Law the gain sits with Qwen 3.5 4B and Gemma 4 E2B, 5.9 and 6.5 points, and Qwen 3.5 2B does not move.

\paragraph{Other retrieved or teacher-written text does not reproduce the gain.} Five instructions drawn at random from the same corpus lower the mean on every task, by 2.0, 2.0, and 0.7 points, so the gain comes from the instructions written for the nearest questions and not from instruction text as such. Few-shot prompting with five worked examples from the same teacher lowers the mean on every task, by 1.6, 0.8, and 0.3 points, so the gain does not come from teacher-written text as such. RICP's principles share the teacher-writes, student-retrieves design and are written from the students' errors. The principles move the mean by $-$0.4, $-$1.1, and 1.2 points, so what the corpus adds is the content written for each cluster. Textbook passages on MedQA, retrieved with the same embedder and the same count, gain 3.9 points against the corpus's 10.6. A passage that holds the fact still leaves the student to find and apply it, where an instruction states the fact in $K_j$ with the procedure $P_j$ and the common mistakes $M_j$ for that kind of question (Section~\ref{sec:told-vs-found}).

\begin{table}[t]
\centering\small
\setlength{\tabcolsep}{4.8pt}
\resizebox{\textwidth}{!}{
\begin{tabular}{@{}l c c c c c c c@{}}
\toprule
 & & & & \multicolumn{4}{c}{Retrieved text} \\
\cmidrule(lr){5-8}
\textbf{Student} & \textbf{Zero-shot} & \textbf{Few-shot} & \textbf{Self-consist.} & \textbf{RICP} & \textbf{Passages} & \textbf{Random instr.} & \textbf{Instructions} \\
\midrule
\rowcolor{bandcol} \multicolumn{8}{@{}l}{\textbf{\textit{MedQA}}} \\
\rowcolor{qwenband} Qwen 3.5 2B & 51.8 & 51.1 & 57.7 & 53.8 & 58.5 & 51.9 & \textbf{67.6} \\
\rowcolor{qwenband} Qwen 3.5 4B & 78.7 & 76.0 & 80.1 & 77.9 & 79.3 & 76.7 & \textbf{84.0} \\
\rowcolor{gemmaband} Gemma 4 E2B & 57.2 & 55.1 & 58.1 & 57.4 & 62.0 & 53.2 & \textbf{69.4} \\
\rowcolor{gemmaband} Gemma 4 E4B & 71.2 & 69.6 & 72.6 & 70.6 & 73.6 & 67.9 & \textbf{80.3} \\
\rowcolor{ministralband} Ministral 3 3B & 62.5 & 61.3 & 64.7 & 59.3 & 67.2 & 61.6 & \textbf{73.1} \\
\rowcolor{meanband} \textbf{\textit{Mean}} & \textbf{64.3} & \textbf{62.6} & \textbf{66.6} & \textbf{63.8} & \textbf{68.1} & \textbf{62.3} & \textbf{74.9} \\
\addlinespace[2pt]
\rowcolor{bandcol} \multicolumn{8}{@{}l}{\textbf{\textit{MMLU Law}}} \\
\rowcolor{qwenband} Qwen 3.5 2B & 40.9 & 41.5 & 40.2 & 36.7 & -- & 40.9 & \textbf{40.9} \\
\rowcolor{qwenband} Qwen 3.5 4B & 56.9 & 55.2 & 60.6 & 58.9 & -- & 59.3 & \textbf{62.8} \\
\rowcolor{gemmaband} Gemma 4 E2B & 46.3 & 46.1 & 48.1 & 48.5 & -- & 45.6 & \textbf{52.8} \\
\rowcolor{gemmaband} Gemma 4 E4B & 60.6 & 57.2 & 59.3 & 58.0 & -- & 54.4 & \textbf{62.2} \\
\rowcolor{ministralband} Ministral 3 3B & 49.8 & 50.2 & 50.7 & 47.0 & -- & 44.3 & \textbf{52.4} \\
\rowcolor{meanband} \textbf{\textit{Mean}} & \textbf{50.9} & \textbf{50.0} & \textbf{51.8} & \textbf{49.8} & -- & \textbf{48.9} & \textbf{54.2} \\
\addlinespace[2pt]
\rowcolor{bandcol} \multicolumn{8}{@{}l}{\textbf{\textit{MathQA}}} \\
\rowcolor{qwenband} Qwen 3.5 2B & 75.7 & 74.5 & 83.2 & 78.8 & -- & 72.8 & \textbf{79.8} \\
\rowcolor{qwenband} Qwen 3.5 4B & 88.3 & 88.6 & 90.4 & 89.1 & -- & 89.2 & \textbf{90.3} \\
\rowcolor{gemmaband} Gemma 4 E2B & 85.8 & 85.6 & 87.5 & 86.4 & -- & 84.3 & \textbf{88.4} \\
\rowcolor{gemmaband} Gemma 4 E4B & 89.0 & 88.3 & 89.7 & 89.9 & -- & 88.9 & \textbf{90.1} \\
\rowcolor{ministralband} Ministral 3 3B & 81.7 & 82.4 & 86.2 & 82.5 & -- & 81.9 & \textbf{84.1} \\
\rowcolor{meanband} \textbf{\textit{Mean}} & \textbf{84.1} & \textbf{83.9} & \textbf{87.4} & \textbf{85.3} & -- & \textbf{83.4} & \textbf{86.5} \\
\bottomrule
\end{tabular}
}
\caption{Test accuracy (\%) of each student under each condition. Bold marks the corpus, the background colour the student family, and the Mean row averages the five students. The last four conditions place retrieved text in the prompt; passages exist for MedQA only; random instructions are five drawn at random from the same corpus; self-consistency votes over three samples where every other condition generates once. On MedQA the corpus raises the mean by 10.6 points, more than any other condition, while random instructions leave it below zero-shot on every task. Intervals per cell are in Appendix~\ref{app:full-results}.}
\label{tab:main}
\end{table}

\paragraph{Self-consistency beats the corpus only on mathematics.} Self-consistency generates three answers where the corpus generates one. On the knowledge-intensive tasks, MedQA and MMLU Law, it gains 2.4 and 0.9 points against the corpus's 10.6 and 3.3. On MathQA it gains 3.3 against the corpus's 2.4 and is the best condition for three of the five students. The students already answer 84.1\% of MathQA test questions zero-shot, and Section~\ref{sec:errors} shows that MathQA has the smallest share of questions a student gets wrong on every attempt; Section~\ref{sec:interventions} explains why a second attempt supplies what an instruction does there. Adding self-consistency to the corpus raises accuracy above the corpus alone on every task (Appendix~\ref{app:full-results}), so the two gains come from different questions.

\paragraph{With the corpus the students close a third of the gap to the frontier reference on MedQA.} The five students' mean closes 35\% of its gap on MedQA (10.6 of 30.4 points), 9\% on MMLU Law, and 41\% on MathQA, where Qwen 3.5 4B and Gemma 4 E4B with the corpus reach 90.3 and 90.1\% and match the reference. Qwen 3.5 4B alone closes 33\% of its gap on MedQA and 18\% on MMLU Law. MMLU Law is the task where the students stand furthest from the reference and where the corpus moves them least; Section~\ref{sec:interventions} returns to why.

\section{Analysis} \label{sec:analysis}

This section asks what kind of wrong answer a small model gives, which intervention and which part of an instruction fixes each kind, and why an instruction beats a passage that holds the same fact.

\subsection{Which kinds of wrong answer do small models give?} \label{sec:errors}

One greedy answer cannot tell a missing fact from an undecided choice, so on the development split we sample five answers per question at temperature 0.7 and classify each question by how many are correct. The classification follows the pass@$k$ framing, in which a problem counts as solved if any of $k$ samples is correct \citep{chen2021codex}, and the share of problems solved at $k$ is called coverage \citep{brown2024monkeys}. A question is \emph{always right} when all five samples are correct, \emph{usually right} when three or four are, \emph{rarely right} when one or two are, and \emph{always wrong} when none is.

\begin{figure}[t]
    \centering
    \includegraphics[width=1\linewidth]{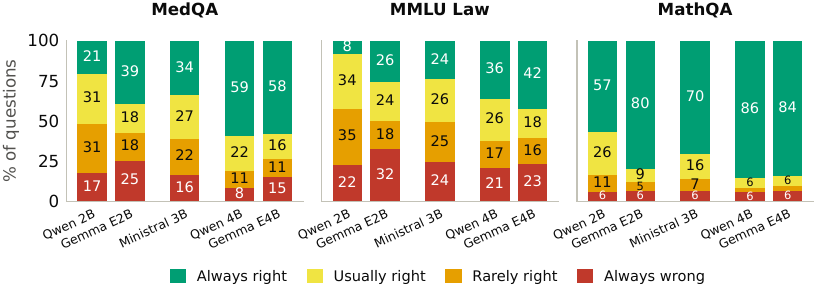}
    \caption{Questions by how many of five sampled answers are correct, for each student and task. Each bar is one student and its segments, from the bottom, are always wrong, rarely right, usually right, and always right. The always-wrong share depends on the task more than on the student, about 6\% on MathQA for every student and between 21\% (Qwen 3.5 4B) and 32\% (Gemma 4 E2B) on MMLU Law.}
    \label{fig:errors}
\end{figure}

\paragraph{The error distribution is driven by task more than model.} A student that is wrong on all five samples is missing a fact, and a student that is sometimes right has what it needs and wavers between options. The share of always-wrong questions is set by the task (Figure~\ref{fig:errors}; per-student shares in Appendix~\ref{app:errors-full}). Every student gets about 6\% of MathQA questions and between a fifth and a third of MMLU Law questions wrong on every sample, whether it is the weakest student or the strongest. MedQA is the one task where a stronger student also reduces that share, from 17\% for Qwen 3.5 2B to 8\% for Qwen 3.5 4B. The toxic-megacolon question of Figure~\ref{fig:corpus} is one every student gets wrong on every attempt, choosing colonoscopy or CT, the two tests the common mistakes warn against.

\subsection{Which intervention fixes which kind of wrong answer?} \label{sec:interventions}

\begin{figure}[t]
    \centering
    \includegraphics[width=1\linewidth]{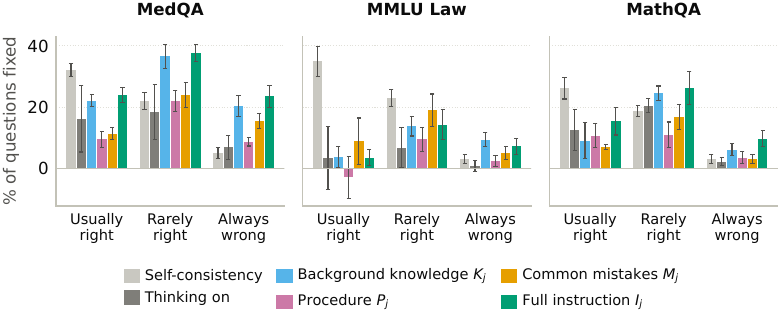}
    \caption{Change in the percent of wrong development questions each intervention fixes, against three fresh samples with nothing added, by kind of wrong answer and task, averaged over the four Qwen and Gemma students; whiskers are the range across the four students. A question is fixed when three fresh samples are all correct. Thinking on and self-consistency fix usually-right and rarely-right questions and almost no always-wrong question. The background knowledge $K_j$ is the only part that reaches the always-wrong questions, and which part fixes most differs by task.}
    \label{fig:interventions}
\end{figure}

If the two kinds of wrong answer have different causes, they should respond to different interventions. We rerun every wrong development question of the four Qwen and Gemma students under six inference-time interventions. Two add deliberation with no new information, thinking on, which these four students have, and self-consistency, the majority vote over three samples. Three add one part of the retrieved instruction $I_j$ on its own, its background knowledge $K_j$, its procedure $P_j$, or its common mistakes $M_j$, and the sixth adds the full $I_j$. A wrong question is \emph{fixed} by an intervention if three fresh samples are all correct under it. Three fresh samples with nothing added already fix some questions, most of the usually-right ones, so Figure~\ref{fig:interventions} reports each intervention's share fixed less the share fixed by that control, averaged over the four students, and Appendix~\ref{app:interventions-full} gives the raw shares.

\paragraph{Background knowledge reaches the always-wrong questions and more deliberation does not.} More samples or a thinking mode let a model reconsider what it already knows, so they fix questions it sometimes gets right and almost none that it always gets wrong. On every task, self-consistency and thinking on each fix under 7\% of the always-wrong questions, while $K_j$ alone fixes 20\% of them on MedQA (Figure~\ref{fig:interventions}). Thinking on helps mainly the Qwen students (Appendix~\ref{app:interventions-full}). Most always-wrong questions stay wrong under every intervention, unless $K_j$ of the retrieved instruction directly states the missing fact. Section~\ref{sec:told-vs-found} measures the case where it does.

\paragraph{Which part of the instruction helps depends on what the task asks for.} A board-exam question is answered by recalling a fact, a bar-exam question by telling apart options that differ in one condition, and a math word problem by carrying out a procedure the student already knows on new numbers. Each task is fixed by the part of the instruction that supplies what it asks for (Figure~\ref{fig:interventions}). On MedQA $K_j$ alone fixes almost as many questions as the full $I_j$, and $P_j$ alone fixes about half as many. On MMLU Law a student that knows the rule can still pick the wrong option, and $M_j$ fixes more rarely-right questions than $K_j$ does, 19\% against 14\%, and more than the full $I_j$. The always-wrong questions are reached only by $K_j$, which fixes a tenth of them. On MathQA no single part fixes as many questions as the full $I_j$, and thinking on fixes as many rarely-right questions as $K_j$ does.

The gains in Table~\ref{tab:main} follow from which part each task needs. On MedQA a stated fact is what a board-exam question needs, so the corpus gains most. On MMLU Law the questions every student gets wrong are the largest share and $K_j$ fixes only a tenth of them, so it gains least. On MathQA a second attempt supplies what the instruction supplies, so self-consistency matches it.

\subsection{Why do instructions beat passages?} \label{sec:told-vs-found}

A retrieved passage and a retrieved instruction can hold the same fact, and on MedQA the passages recover a third of what the instructions recover (Section~\ref{sec:results}). A passage can fall short in two ways. First, it can lack the fact that decides the question. Second, it can hold the fact among other facts that point to a wrong option. On MedQA, the one task with a textbook corpus, GPT-5 as judge reads each of the 1{,}273 test questions with its keyed answer. For the five retrieved passages, and separately for the five retrieved instructions, the judge labels whether the text holds the deciding fact and whether it would push a reader toward a wrong option, following the sufficient-context labelling of prior work \citep[][cf. prompt in Appendix~\ref{prompt-templates}]{joren2025iclr-sufficient}.

\paragraph{Passages hold the fact for most questions and mislead more than twice as often.} Both texts hold the deciding fact for most questions, the instructions for 77\% and the passages for 65\%, and where both hold it the student puts right 58\% of its wrong zero-shot answers from the passages and 65\% from the instructions (mean of the five students). The difference is what surrounds the fact. The judge marks a passage as misleading for 22\% of questions and an instruction for 9\%, and where a text lacks the fact, the passages turn a third of the questions the student had right into wrong answers and the instructions a fifth (Appendix~\ref{app:told-vs-found}). A textbook page carries the fact beside other facts that fit the options, and an instruction written for a cluster of questions carries the fact those questions turn on in $K_j$, what separates the options in $M_j$, and nothing that points elsewhere. The four cells (fact in both texts, in one, in neither) per student are in Appendix~\ref{app:told-vs-found}.

\section{Conclusion}

Distillation compresses a teacher's expertise into a student's weights. Here, we introduce a new approach to compress teacher knowledge into a reusable corpus of instructions that any small instruction-tuned model can benefit from. A teacher writes one instruction per cluster of a domain's problems, background knowledge with a procedure and the common mistakes, once per domain, and frozen small models retrieve and follow them with no training. Only the background knowledge reaches the questions a student gets wrong on every attempt, and the procedure and common mistakes fix only the questions where it wavers, so which part does the work depends on the task. A corpus pays where a student gets a large share of questions wrong on every attempt, as on MedQA and MMLU Law, and where that share is small, as on MathQA, self-consistency is the cheaper remedy. On MedQA the corpus raises the mean accuracy of five small models by 10.6 points and the two 2B models by 12 and 16, from a corpus built once per domain. A changed guideline in medicine or law is then an edit to one instruction, where a fine-tuned model would be retrained.

Future work follows from the findings. Every benchmark here is multiple choice, so whether a student follows an instruction as well on a free-form answer, where no option narrows the choice, is untested. The retriever is an off-the-shelf embedder for instructions and passages alike, so reranking the retrieved instructions or training a retriever on question-to-instruction pairs is untested. Every question retrieves, where cascades escalate only when a scorer judges the answer unreliable \citep{chen2024tmlr-frugalgpt} and adaptive retrieval retrieves only when a signal says the model needs it \citep{mallen2023trust,asai2024iclr-selfrag}, so a student could retrieve only when its confidence or the agreement among its samples says it should. GPT-5 writes every instruction, and whether an open-weight teacher, and how small a one, can write instructions a small model follows is open.

\section*{Reproducibility Statement}

The Instruction Corpus for each domain, every prompt, the retrieval and evaluation code, and every student output behind the tables are in the supplementary material and will be released on publication. Section~\ref{sec:corpus} describes how a corpus is built and Appendix~\ref{app:corpus} gives the prompts, one full instruction, and the development-split sweeps. Section~\ref{sec:setup} fixes the splits, students, baselines, and metrics, and Appendix~\ref{appendix-inference} lists checkpoints and decoding settings. Appendix~\ref{app:full-results} reports every per-student number with bootstrap intervals.

\section*{AI Use Statement}

We used generative AI tools to draft and edit the paper's text, to review its structure and prose, to help implement the method and the evaluation and figure scripts, and, in part, to interpret results and design experiments. We did not use them to generate data, to formulate or prove claims, or to clean datasets. Every number in the paper was checked against the tables printed by our own code, and all AI-assisted text was reviewed by the authors, who take responsibility for the final content. Separately, a large language model is a component of the system under study: GPT-5 wrote the Instruction Corpus, the few-shot exemplars, and the principles of the RICP baseline, and labelled whether a retrieved passage or instruction holds the fact a question needs (Section~\ref{sec:told-vs-found}).

\bibliography{references}
\bibliographystyle{iclr2027_conference}

\clearpage
\appendix
\raggedbottom
\renewcommand{\floatpagefraction}{0.75}
\setcounter{topnumber}{3}

\section{Corpus Details} \label{app:corpus}

This appendix gives the prompts that built the corpus and ran the students, one full instruction with the question it decided, and the development-split sweeps that fixed the cluster size and the number of instructions retrieved.

\subsection{Prompts} \label{prompt-templates}

\paragraph{The teacher's instruction request} One request per cluster; the domain is ``medical'', ``law'', or ``mathematics''. The problems block lists the cluster's training questions in the form below, and under a question the two wrong options the students chose most, each with the end of one attempt that led there, pooled over the four Qwen and Gemma students with four attempts each. A cluster on which no student erred gets the request without the mistakes sentence and with the ask ``What are the most common mistakes that lead someone to a wrong answer on problems of this type?''. The reply's three lists are the instruction's three parts; the paper names \texttt{application\_steps} the procedure $P_j$ and \texttt{discriminating\_features} the common mistakes $M_j$.

\begin{lstlisting}
Here are {n} {domain} problems of one type, each with its options and correct answer. Under some of them are wrong answers that students gave, with the end of the reasoning that led there.

Write three things for someone who will meet a new problem of this type. Keep each one short: say what matters most and leave out the rest.
Do not restate any of these problems or their answers.

"background_knowledge": What background knowledge is most important for solving problems on this topic? Cover the topic, not only these problems: include the closely related conditions, rules or methods that a new problem on this topic could turn on, and what tells them apart. State each fact directly.
"application_steps": What are the most important steps for reasoning about problems of this type?
"discriminating_features": What mistakes did the students make on these problems, and what separates the correct answer from the one they chose? Write it as guidance for any problem of this type.

Also give "tests": one line naming what problems of this type test.

Return only a JSON object:
{"tests": "<one line>", "background_knowledge": ["<sentence>", ...], "application_steps": ["<sentence>", ...], "discriminating_features": ["<sentence>", ...]}

<problems>
{problems}
</problems>
\end{lstlisting}

One problem inside the request:

\begin{lstlisting}
Problem: {question}
Options:
{options, one per line as "A. text"}
Correct answer: {letter}. {text}
Students chose {letter}. {text} in {n} of 16 attempts. The end of one such attempt: "... {last 60 words}"
\end{lstlisting}

\paragraph{The student prompt} Every condition places its text under one header, then the question and the output format of the task's zero-shot prompt; zero-shot omits the header block. Instructions, passages, RICP principles, worked examples, and random instructions are separated by a rule. An instruction is rendered as its three parts under their names, one sentence per line, and the one-part interventions of Section~\ref{sec:interventions} render that part alone. MedQA's format block is shown; MMLU Law and MathQA differ only in the wording of the reasoning field.

\begin{lstlisting}
# Reference material. The following may or may not apply to this question; use what does and ignore the rest.

{text 1}

---

{text 2}

...

# Question
{question with its options}

# Required Output Format
```json
{
  "reasoning": "<step-by-step analysis of the medical scenario and answer choices>",
  "final_answer": "<letter of the chosen answer choice>"
}
```
\end{lstlisting}

\paragraph{Few-shot exemplars} The teacher writes one worked solution per training question with the request below. The five exemplars of each task, the same five for every question, are in the released code.

\begin{lstlisting}
Here is a {domain} multiple-choice question and its correct answer.

Write a worked solution a student could learn from: reason step by step from the question to the answer, in one short paragraph.

Return only a JSON object: {"reasoning": "<the worked solution>"}

<question>
{question}
</question>

<options>
{options}
</options>

Correct answer: {gold}
\end{lstlisting}

\paragraph{RICP} The RICP baseline's two teacher requests, as published. One insight request per wrong greedy training attempt of the student, then one principle request per cluster of the reasons. The large setting is run, ten task-level principles above every question and ten question-level principles retrieved by the question's nearest mistakes.

\begin{lstlisting}
A student model answered the following {domain} problem incorrectly. Its reasoning and the correct answer are below.

Give the high-level reason for the mistake, as a short category (a few words), and a few insights: specific guidance that would help the student avoid the same kind of mistake on a similar problem.
Do not mention this problem's options or answer.

Return only a JSON object: {"reason": "<a few words>", "insights": ["<sentence>", ...]}

{error}
\end{lstlisting}

\begin{lstlisting}
Below are {n} mistakes a student model made on {domain} problems, each with the reason for it. They share one kind of mistake. Write one task-level principle the student should follow to avoid this kind of mistake on any {domain} problem: a short paragraph, general, mentioning no problem below.

Return only a JSON object: {"principle": "<the principle>"}

<mistakes>
{mistakes}
</mistakes>
\end{lstlisting}

\paragraph{The judge} Sent once per test question and text in Section~\ref{sec:told-vs-found}, with the five retrieved passages or the five retrieved instructions as the reference; a ``yes'' counts as holding the fact.

\begin{lstlisting}
Here is a {domain} multiple-choice question with its correct answer, and a reference text.

Does the reference text hold the knowledge needed to decide this question?
- "yes": it states the deciding point.
- "partly": it states something that helps, but not the deciding point.
- "no": it is about something else.
Also say whether the reference text would push a reader toward a wrong option.

Return only a JSON object: {"holds": "yes" | "partly" | "no", "misleads": true | false}

<question>
{question}
</question>

Correct answer: {gold}

<reference>
{reference}
</reference>
\end{lstlisting}

\subsection{A Full Instruction} \label{app:instruction}

Instruction g1773 is the first of the five the toxic-megacolon question of Figure~\ref{fig:corpus} retrieves (test question medqa\_000320, keyed answer D; the five retrieved are g1773, g3304, g1088, g0900, and g0427). The teacher wrote it for three training questions, and its line on what the cluster tests reads ``Recognition and initial management of suspected toxic megacolon in ulcerative colitis flares (first-line imaging, treatment priorities, and contraindications).'' The instruction follows, then the question and Qwen 3.5 2B's answer with the corpus; zero-shot it answers colonoscopy or CT on every attempt.

\begin{lstlisting}
Background knowledge
Toxic megacolon = systemic toxicity plus nonobstructive colonic dilation (>6 cm), best seen on plain abdominal x-ray.
Suspect toxic megacolon in IBD with fever, tachycardia, hypotension, severe abdominal distension/pain, or peritoneal signs.
First-line imaging when megacolon or perforation is possible is a plain abdominal radiograph (KUB); look for dilation and free air.
Colonoscopy and contrast enemas are contraindicated in suspected toxic megacolon or perforation due to perforation risk.
CT can help define complications but is not the first step when KUB can rapidly confirm dilation or free air, especially in unstable patients.
Acute severe UC without megacolon: admit, bowel rest, IV fluids, IV corticosteroids (e.g., hydrocortisone or methylprednisolone), correct electrolytes, DVT prophylaxis, test for C. difficile.
Confirmed toxic megacolon from IBD: bowel rest, NG decompression, IV fluids, IV corticosteroids, broad-spectrum IV antibiotics, early surgical consult.
Fulminant C. difficile colitis (not UC flare) requires high-dose oral/NG vancomycin plus IV metronidazole; consider rectal vancomycin if ileus.
Operate urgently for perforation, uncontrolled bleeding, or failure of medical therapy (typically within 24–72 hours).
Avoid antidiarrheals, opioids, and bowel preps in severe colitis or suspected megacolon.

Procedure
Identify red flags for toxic megacolon or perforation (systemic toxicity, distension, severe pain, peritoneal signs, instability).
Obtain a plain abdominal x-ray immediately to assess colonic diameter and look for free air; avoid colonoscopy/contrast studies.
If dilation without perforation: initiate bowel rest, IV fluids, electrolyte correction, IV corticosteroids (if UC), IV antibiotics, NG decompression, and surgical consult.
If no megacolon and severe UC flare: treat with IV corticosteroids and supportive care; rule out infections (e.g., C. difficile).
Escalate to CT only if diagnosis remains unclear after x-ray or to define complications in a stable patient.
Proceed to surgery for perforation, peritonitis, uncontrolled hemorrhage, or failure of medical therapy despite optimal care.

Common mistakes
Do not start oral/topical therapy in severe disease; severe/fulminant UC needs IV corticosteroids, not oral prednisone or rectal 5-ASA.
Choose a plain abdominal x-ray first to confirm dilation or free air; CT is not the initial test when megacolon is suspected and time/safety matter.
Avoid colonoscopy and contrast enemas in suspected megacolon; they increase perforation risk and do not change immediate management.
Do not jump to colectomy without imaging confirmation unless there is unequivocal perforation; stabilize and image quickly.
Reserve metronidazole/vancomycin regimens for fulminant C. difficile, not for UC flares without evidence of infection.
\end{lstlisting}

Test question medqa\_000320 (keyed answer D):

\begin{lstlisting}
A 32-year-old woman comes to the emergency department because a 5-week history of abdominal pain and bloody diarrhea that has worsened in the past 24 hours. She was diagnosed with ulcerative colitis 1 year ago but has had difficulty complying with her drug regimen. Her temperature is 38.2°C (100.8°F), pulse is 120/min, and blood pressure is 92/56 mm Hg. Examination shows a distended, rigid abdomen and hypoactive bowel sounds. Fluid resuscitation is initiated. In addition to complete bowel rest, which of the following is the most appropriate next step in management of this patient?

A. Double-contrast barium enema
B. Abdominal CT scan with contrast
C. CT angiography
D. Abdominal x-ray
E. Colonoscopy
\end{lstlisting}

Qwen 3.5 2B with the corpus, answer D:

\begin{lstlisting}
The patient presents with classic signs of toxic megacolon: fever, tachycardia, hypotension, severe abdominal distension, rigidity, and peritoneal signs. The history of ulcerative colitis (UC) and non-compliance with medication increases the risk of flare and progression to megacolon. According to the provided reference material, the first-line imaging when megacolon or perforation is possible is a plain abdominal radiograph (KUB) to look for colonic dilation and free air. Colonoscopy and contrast enemas are explicitly contraindicated due to the high risk of perforation in this setting. While CT with contrast is useful for defining complications, it is not the first step when a plain x-ray can rapidly confirm the diagnosis (dilation) or free air, especially in an unstable patient. Therefore, the immediate next step is to obtain a plain abdominal x-ray to confirm the diagnosis before proceeding to more invasive procedures like colonoscopy or CT.
\end{lstlisting}

\subsection{Development-Split Selections} \label{app:selection}

The cluster size and the number of instructions retrieved ($k$) are chosen on the selection half of the development split (636 MedQA and 85 MMLU Law questions), greedy, over the four Qwen and Gemma students. Table~\ref{tab:sweep} gives the gain over zero-shot in points, mean of the four, for each cluster size and $k$. Size 3 at $k=5$ is chosen. MMLU Law's 85 questions cannot separate the sizes and take MedQA's setting, as does MathQA. On the held-out half of the development split the chosen setting gains 11.2 points on MedQA against 9.9 on the selection half.

\begin{table}[htbp]
\centering\small

\begin{tabular}{@{}l c c c c c@{}}
\toprule
\textbf{Task} & \textbf{k} & \textbf{Size 1} & \textbf{Size 3} & \textbf{Size 8} & \textbf{Size 27} \\
\midrule
MedQA & 1 & +5.7 & +4.0 & +5.5 & +0.8 \\
MedQA & 3 & +9.5 & +8.8 & +6.6 & +4.0 \\
MedQA & 5 & +11.5 & +11.0 & +9.1 & -- \\
MMLU Law & 1 & +5.3 & $-$0.6 & +1.2 & +0.3 \\
MMLU Law & 3 & +6.5 & +0.6 & $-$0.3 & +0.0 \\
MMLU Law & 5 & +5.3 & +4.7 & +5.6 & -- \\
\bottomrule
\end{tabular}
\caption{Gain over zero-shot in accuracy points (mean of the four Qwen and Gemma students) on the selection half of the development split, by cluster size (questions per cluster) and number of instructions retrieved $k$. The chosen cell is size 3 at $k=5$; size 27 was not run at $k=5$.}
\label{tab:sweep}
\end{table}

\FloatBarrier

\section{Inference Settings} \label{appendix-inference}

Every run uses vLLM on one GPU in half precision. Greedy decoding is temperature 0; the three samples of self-consistency and of every sampled run are at temperature 0.7 and top-p 0.95 for every student. Thinking mode is off everywhere except the thinking-on intervention of Section~\ref{sec:interventions}, which turns it on for the Qwen and Gemma students at the caps in Table~\ref{tab:checkpoints}. The teacher, the frontier reference, and the judge of Section~\ref{sec:told-vs-found} are all \texttt{gpt-5-2025-08-07} through the OpenAI API at its default settings, including the default reasoning effort, for all three roles. The answer is read from the first complete JSON object in the reply (\texttt{final\_answer}); if none parses, from a malformed JSON field, then from the last ``answer: X'' or parenthetical letter in the prose, and on MathQA a reply that states a value and no letter is matched to the option holding that value. A reply that yields no answer is a format failure and is scored wrong; nearly all are replies cut at the generation cap by a student that never states an answer.

\begin{table}[htbp]
\centering\footnotesize
\setlength{\tabcolsep}{4pt}
\begin{tabular}{@{}l l r r r@{}}
\toprule
\textbf{Student} & \textbf{Checkpoint} & \textbf{Context} & \textbf{Generation cap} & \textbf{Thinking cap} \\
\midrule
Qwen 3.5 2B & {\scriptsize \texttt{Qwen/Qwen3.5-2B}} & 32,768 & 10,000 & 16,384 \\
Qwen 3.5 4B & {\scriptsize \texttt{Qwen/Qwen3.5-4B}} & 32,768 & 10,000 & 16,384 \\
Gemma 4 E2B & {\scriptsize \texttt{google/gemma-4-E2B-it}} & 16,384 & 8,192 & 12,288 \\
Gemma 4 E4B & {\scriptsize \texttt{google/gemma-4-E4B-it}} & 16,384 & 8,192 & 12,288 \\
Ministral 3 3B & {\scriptsize \texttt{mistralai/Ministral-3-3B-Instruct-2512}} & 32,768 & 10,000 & -- \\
\bottomrule
\end{tabular}
\caption{Checkpoints, context windows, and generation and thinking caps (tokens) of every student; Ministral 3 3B has no thinking mode. Every sampled run uses temperature 0.7 and top-p 0.95.}
\label{tab:checkpoints}
\end{table}

\FloatBarrier

\section{Full Results} \label{app:full-results}

This appendix carries the per-student numbers behind every exhibit in Sections~\ref{sec:results} and~\ref{sec:analysis}.

\subsection{Main Results} \label{app:main-full}

Table~\ref{tab:full-results} gives test accuracy under every condition of Table~\ref{tab:main} with the 95\% bootstrap interval of the paired difference from zero-shot over test questions, and adds the instructions-plus-self-consistency condition. Table~\ref{tab:format-failures} counts the format failures per cell.

\begin{table}[htbp]
\centering\scriptsize
\setlength{\tabcolsep}{3pt}
\begin{tabular}{@{}l c c c c c c@{}}
\toprule
\textbf{Condition} & \textbf{Qwen 3.5 2B} & \textbf{Qwen 3.5 4B} & \textbf{Gemma 4 E2B} & \textbf{Gemma 4 E4B} & \textbf{Ministral 3 3B} & \textbf{Mean of five} \\
\midrule
\rowcolor{bandcol} \multicolumn{7}{@{}l}{\textbf{\textit{MedQA}}} \\
Zero-shot & 51.8 & 78.7 & 57.2 & 71.2 & 62.5 & 64.3 \\
Few-shot & \ci{51.1}{[$-$3.7, $+$2.3]} & \ci{76.0}{[$-$4.9, $-$0.5]} & \ci{55.1}{[$-$4.5, $+$0.5]} & \ci{69.6}{[$-$3.8, $+$0.5]} & \ci{61.3}{[$-$3.7, $+$1.4]} & 62.6 ($-$1.6) \\
Self-consist. & \ci{57.7}{[$+$3.1, $+$8.8]} & \ci{80.1}{[$-$0.6, $+$3.5]} & \ci{58.1}{[$-$0.9, $+$2.7]} & \ci{72.6}{[$-$0.2, $+$3.1]} & \ci{64.7}{[$-$0.4, $+$4.8]} & 66.6 ($+$2.4) \\
RICP & \ci{53.8}{[$-$1.1, $+$5.2]} & \ci{77.9}{[$-$3.0, $+$1.3]} & \ci{57.4}{[$-$2.4, $+$2.7]} & \ci{70.6}{[$-$2.7, $+$1.5]} & \ci{59.3}{[$-$6.0, $-$0.3]} & 63.8 ($-$0.4) \\
Passages & \ci{58.5}{[$+$3.6, $+$10.0]} & \ci{79.3}{[$-$1.8, $+$2.8]} & \ci{62.0}{[$+$2.1, $+$7.5]} & \ci{73.6}{[$+$0.2, $+$4.9]} & \ci{67.2}{[$+$2.2, $+$7.5]} & 68.1 ($+$3.9) \\
Random instr. & \ci{51.9}{[$-$2.8, $+$3.1]} & \ci{76.7}{[$-$4.0, $+$0.1]} & \ci{53.2}{[$-$6.6, $-$1.4]} & \ci{67.9}{[$-$5.3, $-$1.3]} & \ci{61.6}{[$-$3.5, $+$1.7]} & 62.3 ($-$2.0) \\
Instructions & \ci{67.6}{[$+$12.7, $+$19.0]} & \ci{84.0}{[$+$3.1, $+$7.5]} & \ci{69.4}{[$+$9.3, $+$15.0]} & \ci{80.3}{[$+$6.8, $+$11.5]} & \ci{73.1}{[$+$7.9, $+$13.3]} & 74.9 ($+$10.6) \\
Instr. $+$ self-consist. & \ci{71.2}{[$+$16.5, $+$22.5]} & \ci{84.9}{[$+$3.9, $+$8.4]} & \ci{70.4}{[$+$10.4, $+$16.2]} & \ci{79.3}{[$+$5.9, $+$10.5]} & \ci{76.1}{[$+$10.9, $+$16.4]} & 76.4 ($+$12.2) \\
\addlinespace[2pt]
\rowcolor{bandcol} \multicolumn{7}{@{}l}{\textbf{\textit{MMLU Law}}} \\
Zero-shot & 40.9 & 56.9 & 46.3 & 60.6 & 49.8 & 50.9 \\
Few-shot & \ci{41.5}{[$-$3.9, $+$5.6]} & \ci{55.2}{[$-$5.6, $+$2.6]} & \ci{46.1}{[$-$4.3, $+$3.9]} & \ci{57.2}{[$-$7.2, $+$0.6]} & \ci{50.2}{[$-$4.3, $+$5.0]} & 50.0 ($-$0.8) \\
Self-consist. & \ci{40.2}{[$-$5.6, $+$3.7]} & \ci{60.6}{[$+$0.0, $+$7.4]} & \ci{48.1}{[$-$1.9, $+$5.7]} & \ci{59.3}{[$-$4.6, $+$2.0]} & \ci{50.7}{[$-$3.2, $+$5.0]} & 51.8 ($+$0.9) \\
RICP & \ci{36.7}{[$-$9.3, $+$0.6]} & \ci{58.9}{[$-$2.0, $+$6.3]} & \ci{48.5}{[$-$2.4, $+$7.0]} & \ci{58.0}{[$-$6.7, $+$1.7]} & \ci{47.0}{[$-$7.6, $+$2.0]} & 49.8 ($-$1.1) \\
Random instr. & \ci{40.9}{[$-$5.0, $+$4.6]} & \ci{59.3}{[$-$1.9, $+$6.7]} & \ci{45.6}{[$-$5.4, $+$3.9]} & \ci{54.4}{[$-$9.8, $-$2.4]} & \ci{44.3}{[$-$10.0, $-$1.1]} & 48.9 ($-$2.0) \\
Instructions & \ci{40.9}{[$-$5.4, $+$5.2]} & \ci{62.8}{[$+$1.7, $+$10.6]} & \ci{52.8}{[$+$2.0, $+$11.1]} & \ci{62.2}{[$-$2.8, $+$6.1]} & \ci{52.4}{[$-$2.2, $+$7.2]} & 54.2 ($+$3.3) \\
Instr. $+$ self-consist. & \ci{46.1}{[$+$0.2, $+$10.2]} & \ci{63.5}{[$+$2.4, $+$10.9]} & \ci{51.5}{[$+$0.9, $+$9.4]} & \ci{64.3}{[$-$0.6, $+$8.0]} & \ci{52.4}{[$-$2.4, $+$7.4]} & 55.6 ($+$4.7) \\
\addlinespace[2pt]
\rowcolor{bandcol} \multicolumn{7}{@{}l}{\textbf{\textit{MathQA}}} \\
Zero-shot & 75.7 & 88.3 & 85.8 & 89.0 & 81.7 & 84.1 \\
Few-shot & \ci{74.5}{[$-$2.9, $+$0.3]} & \ci{88.6}{[$-$0.6, $+$1.2]} & \ci{85.6}{[$-$1.2, $+$0.9]} & \ci{88.3}{[$-$1.6, $+$0.1]} & \ci{82.4}{[$-$0.6, $+$1.8]} & 83.9 ($-$0.3) \\
Self-consist. & \ci{83.2}{[$+$6.0, $+$8.8]} & \ci{90.4}{[$+$1.3, $+$2.9]} & \ci{87.5}{[$+$0.8, $+$2.5]} & \ci{89.7}{[$+$0.0, $+$1.4]} & \ci{86.2}{[$+$3.4, $+$5.6]} & 87.4 ($+$3.3) \\
RICP & \ci{78.8}{[$+$1.5, $+$4.5]} & \ci{89.1}{[$-$0.1, $+$1.7]} & \ci{86.4}{[$-$0.5, $+$1.6]} & \ci{89.9}{[$+$0.1, $+$1.6]} & \ci{82.5}{[$-$0.5, $+$1.9]} & 85.3 ($+$1.2) \\
Random instr. & \ci{72.8}{[$-$4.5, $-$1.4]} & \ci{89.2}{[$+$0.0, $+$1.7]} & \ci{84.3}{[$-$2.6, $-$0.4]} & \ci{88.9}{[$-$0.9, $+$0.7]} & \ci{81.9}{[$-$1.0, $+$1.4]} & 83.4 ($-$0.7) \\
Instructions & \ci{79.8}{[$+$2.4, $+$5.6]} & \ci{90.3}{[$+$1.1, $+$3.0]} & \ci{88.4}{[$+$1.5, $+$3.7]} & \ci{90.1}{[$+$0.2, $+$1.8]} & \ci{84.1}{[$+$1.0, $+$3.6]} & 86.5 ($+$2.4) \\
Instr. $+$ self-consist. & \ci{85.7}{[$+$8.4, $+$11.5]} & \ci{91.2}{[$+$2.0, $+$3.7]} & \ci{89.9}{[$+$3.1, $+$5.1]} & \ci{91.2}{[$+$1.3, $+$2.9]} & \ci{87.7}{[$+$4.7, $+$7.2]} & 89.1 ($+$5.0) \\
\bottomrule
\end{tabular}
\caption{Test accuracy (\%) per student under every condition, with the 95\% bootstrap interval of the paired difference from zero-shot beneath each value; the last column is the mean of the five students with its difference from zero-shot. Instructions $+$ self-consistency is the vote over the instruction run's three samples.}
\label{tab:full-results}
\end{table}

\begin{table}[htbp]
\centering\small
\begin{tabular}{@{}l r r r r r r@{}}
\toprule
\textbf{Student} & \textbf{Zero-shot} & \textbf{Few-shot} & \textbf{RICP} & \textbf{Passages} & \textbf{Random instructions} & \textbf{Instructions} \\
\midrule
\rowcolor{bandcol} \multicolumn{7}{@{}l}{\textbf{\textit{MedQA}}} \\
Qwen 3.5 2B & 3 & 2 & 11 & 16 & 10 & 18 \\
Qwen 3.5 4B & 1 & 0 & 0 & 0 & 2 & 0 \\
Gemma 4 E2B & 1 & 0 & 8 & 2 & 1 & 6 \\
Gemma 4 E4B & 1 & 0 & 0 & 2 & 0 & 1 \\
Ministral 3 3B & 4 & 10 & 16 & 14 & 11 & 6 \\
\addlinespace[2pt]
\rowcolor{bandcol} \multicolumn{7}{@{}l}{\textbf{\textit{MMLU Law}}} \\
Qwen 3.5 2B & 8 & 2 & 14 & -- & 10 & 9 \\
Qwen 3.5 4B & 0 & 1 & 0 & -- & 0 & 0 \\
Gemma 4 E2B & 3 & 0 & 0 & -- & 5 & 3 \\
Gemma 4 E4B & 0 & 0 & 0 & -- & 0 & 0 \\
Ministral 3 3B & 6 & 1 & 3 & -- & 4 & 4 \\
\addlinespace[2pt]
\rowcolor{bandcol} \multicolumn{7}{@{}l}{\textbf{\textit{MathQA}}} \\
Qwen 3.5 2B & 255 & 204 & 210 & -- & 355 & 231 \\
Qwen 3.5 4B & 37 & 31 & 55 & -- & 45 & 31 \\
Gemma 4 E2B & 29 & 14 & 14 & -- & 17 & 20 \\
Gemma 4 E4B & 5 & 8 & 7 & -- & 3 & 9 \\
Ministral 3 3B & 172 & 139 & 153 & -- & 152 & 99 \\
\bottomrule
\end{tabular}
\caption{Format failures per cell: greedy replies from which no answer could be read, scored wrong. Passages exist for MedQA only.}
\label{tab:format-failures}
\end{table}

\FloatBarrier

\subsection{Error Categories} \label{app:errors-full}

Table~\ref{tab:errors-full} gives Figure~\ref{fig:errors} per student, the percent of development questions in each category by how many of five sampled zero-shot answers are correct.

\begin{table}[htbp]
\centering\small
\begin{tabular}{@{}l r r r r r@{}}
\toprule
\textbf{Student} & \textbf{n} & \textbf{Always right} & \textbf{Usually right} & \textbf{Rarely right} & \textbf{Always wrong} \\
\midrule
\rowcolor{bandcol} \multicolumn{6}{@{}l}{\textbf{\textit{MedQA}}} \\
Qwen 3.5 2B & 1,272 & 20.9 & 30.8 & 31.0 & 17.3 \\
Qwen 3.5 4B & 1,272 & 59.4 & 21.8 & 10.9 & 7.9 \\
Gemma 4 E2B & 1,272 & 39.3 & 18.0 & 17.5 & 25.2 \\
Gemma 4 E4B & 1,272 & 57.9 & 16.0 & 11.2 & 14.8 \\
Ministral 3 3B & 1,272 & 34.1 & 27.1 & 22.5 & 16.3 \\
\addlinespace[2pt]
\rowcolor{bandcol} \multicolumn{6}{@{}l}{\textbf{\textit{MMLU Law}}} \\
Qwen 3.5 2B & 170 & 8.2 & 34.1 & 35.3 & 22.4 \\
Qwen 3.5 4B & 170 & 36.5 & 25.9 & 17.1 & 20.6 \\
Gemma 4 E2B & 170 & 25.9 & 24.1 & 17.6 & 32.4 \\
Gemma 4 E4B & 170 & 42.4 & 18.2 & 16.5 & 22.9 \\
Ministral 3 3B & 170 & 24.1 & 26.5 & 25.3 & 24.1 \\
\addlinespace[2pt]
\rowcolor{bandcol} \multicolumn{6}{@{}l}{\textbf{\textit{MathQA}}} \\
Qwen 3.5 2B & 1,000 & 56.9 & 26.5 & 10.6 & 6.0 \\
Qwen 3.5 4B & 1,000 & 85.5 & 6.2 & 2.6 & 5.7 \\
Gemma 4 E2B & 1,000 & 79.7 & 8.6 & 5.4 & 6.3 \\
Gemma 4 E4B & 1,000 & 84.4 & 6.1 & 3.1 & 6.4 \\
Ministral 3 3B & 1,000 & 70.4 & 15.9 & 7.3 & 6.4 \\
\bottomrule
\end{tabular}
\caption{Percent of development questions per category: always right (five of five samples correct), usually right (three or four), rarely right (one or two), always wrong (none).}
\label{tab:errors-full}
\end{table}

\FloatBarrier

\subsection{Interventions} \label{app:interventions-full}

Table~\ref{tab:interventions-full} gives Figure~\ref{fig:interventions} per student. For each category of wrong development question it gives the raw percent an intervention fixes (three fresh samples all correct), before the nothing-added control that Figure~\ref{fig:interventions} subtracts. Ministral 3 3B has no intervention runs.

\begin{table}[htbp]
\centering\scriptsize
\setlength{\tabcolsep}{3pt}

\begin{tabular}{@{}l l r r r r r r@{}}
\toprule
\textbf{Student} & \textbf{Category} & \textbf{Self-consist.} & \textbf{Thinking on} & \textbf{Background} & \textbf{Procedure} & \textbf{Mistakes} & \textbf{Full instr.} \\
\midrule
\rowcolor{bandcol} \multicolumn{8}{@{}l}{\textbf{\textit{MedQA}}} \\
Qwen 3.5 2B & Usually (392) & 70.4 & 65.3 & 59.2 & 45.7 & 44.4 & 61.2 \\
Qwen 3.5 2B & Rarely (394) & 24.1 & 25.1 & 30.5 & 17.8 & 17.5 & 35.5 \\
Qwen 3.5 2B & Always wrong (220) & 10.0 & 10.5 & 18.6 & 5.9 & 11.8 & 21.4 \\
Qwen 3.5 4B & Usually (277) & 78.3 & 84.8 & 71.8 & 60.3 & 62.1 & 75.1 \\
Qwen 3.5 4B & Rarely (139) & 38.8 & 51.1 & 47.5 & 34.5 & 40.3 & 51.8 \\
Qwen 3.5 4B & Always wrong (101) & 8.9 & 18.8 & 13.9 & 10.9 & 13.9 & 16.8 \\
Gemma 4 E2B & Usually (229) & 70.7 & 42.4 & 62.0 & 46.7 & 52.0 & 63.3 \\
Gemma 4 E2B & Rarely (223) & 22.9 & 13.9 & 43.0 & 25.6 & 30.9 & 43.9 \\
Gemma 4 E2B & Always wrong (320) & 1.6 & 0.9 & 23.1 & 11.9 & 17.2 & 27.8 \\
Gemma 4 E4B & Usually (204) & 80.9 & 44.6 & 67.2 & 57.4 & 58.8 & 68.6 \\
Gemma 4 E4B & Rarely (143) & 32.2 & 13.3 & 55.2 & 40.6 & 37.1 & 49.7 \\
Gemma 4 E4B & Always wrong (188) & 2.7 & 0.5 & 28.7 & 9.0 & 22.3 & 31.4 \\
\addlinespace[2pt]
\rowcolor{bandcol} \multicolumn{8}{@{}l}{\textbf{\textit{MMLU Law}}} \\
Qwen 3.5 2B & Usually (58) & 77.6 & 37.9 & 36.2 & 22.4 & 36.2 & 34.5 \\
Qwen 3.5 2B & Rarely (60) & 35.0 & 15.0 & 11.7 & 5.0 & 8.3 & 10.0 \\
Qwen 3.5 2B & Always wrong (38) & 0.0 & 0.0 & 7.9 & 0.0 & 0.0 & 0.0 \\
Qwen 3.5 4B & Usually (44) & 68.2 & 59.1 & 43.2 & 47.7 & 61.4 & 40.9 \\
Qwen 3.5 4B & Rarely (29) & 27.6 & 27.6 & 13.8 & 10.3 & 24.1 & 27.6 \\
Qwen 3.5 4B & Always wrong (35) & 5.7 & 5.7 & 5.7 & 2.9 & 2.9 & 8.6 \\
Gemma 4 E2B & Usually (41) & 80.5 & 48.8 & 41.5 & 43.9 & 53.7 & 51.2 \\
Gemma 4 E2B & Rarely (30) & 26.7 & 6.7 & 30.0 & 26.7 & 36.7 & 16.7 \\
Gemma 4 E2B & Always wrong (55) & 1.8 & 0.0 & 16.4 & 7.3 & 9.1 & 12.7 \\
Gemma 4 E4B & Usually (31) & 80.6 & 35.5 & 61.3 & 41.9 & 51.6 & 54.8 \\
Gemma 4 E4B & Rarely (28) & 25.0 & 0.0 & 21.4 & 17.9 & 28.6 & 25.0 \\
Gemma 4 E4B & Always wrong (39) & 7.7 & 0.0 & 10.3 & 2.6 & 10.3 & 10.3 \\
\addlinespace[2pt]
\rowcolor{bandcol} \multicolumn{8}{@{}l}{\textbf{\textit{MathQA}}} \\
Qwen 3.5 2B & Usually (265) & 85.3 & 76.2 & 60.8 & 54.3 & 58.9 & 72.8 \\
Qwen 3.5 2B & Rarely (106) & 36.8 & 32.1 & 31.1 & 19.8 & 20.8 & 28.3 \\
Qwen 3.5 2B & Always wrong (60) & 6.7 & 6.7 & 5.0 & 3.3 & 3.3 & 11.7 \\
Qwen 3.5 4B & Usually (62) & 79.0 & 80.6 & 59.7 & 71.0 & 67.7 & 64.5 \\
Qwen 3.5 4B & Rarely (26) & 23.1 & 34.6 & 34.6 & 11.5 & 34.6 & 26.9 \\
Qwen 3.5 4B & Always wrong (57) & 7.0 & 5.3 & 5.3 & 1.8 & 1.8 & 7.0 \\
Gemma 4 E2B & Usually (86) & 83.7 & 62.8 & 79.1 & 74.4 & 58.1 & 77.9 \\
Gemma 4 E2B & Rarely (54) & 24.1 & 24.1 & 31.5 & 18.5 & 24.1 & 42.6 \\
Gemma 4 E2B & Always wrong (63) & 1.6 & 0.0 & 11.1 & 9.5 & 6.3 & 17.5 \\
Gemma 4 E4B & Usually (61) & 86.9 & 60.7 & 67.2 & 73.8 & 73.8 & 77.0 \\
Gemma 4 E4B & Rarely (31) & 32.3 & 32.3 & 41.9 & 35.5 & 29.0 & 48.4 \\
Gemma 4 E4B & Always wrong (64) & 0.0 & 0.0 & 6.2 & 3.1 & 4.7 & 6.2 \\
\bottomrule
\end{tabular}

\caption{Percent of wrong development questions each intervention fixes, per student and category (with the category's size); the columns are the six interventions of Figure~\ref{fig:interventions}. Intervals per cell are in the released results.}
\label{tab:interventions-full}
\end{table}

\FloatBarrier

\subsection{Fact Told Against Fact Found} \label{app:told-vs-found}

The judge of Section~\ref{sec:told-vs-found} reads each of the 1{,}273 MedQA test questions with its keyed answer against the five retrieved passages and, separately, the five retrieved instructions. Over the 1{,}273 questions the judge marks the passages as holding the fact for 821 (partly for 280, not at all for 172) and as misleading for 285, and the instructions as holding it for 982 (partly 194, not at all 97) and as misleading for 112. Table~\ref{tab:told-vs-found} groups the questions by which text holds the fact (``partly'' counts as not holding it) and splits them by the student's greedy zero-shot answer; each cell gives the group's size and the percent of it the passages and the instructions put right, or, for questions right zero-shot, put wrong, on the greedy test runs of Table~\ref{tab:main}.

\begin{table}[htbp]
\centering\scriptsize
\setlength{\tabcolsep}{3pt}

\begin{tabular}{@{}l l l l l l l@{}}
\toprule
\textbf{Fact held by} & \textbf{Qwen 3.5 2B} & \textbf{Qwen 3.5 4B} & \textbf{Gemma 4 E2B} & \textbf{Gemma 4 E4B} & \textbf{Ministral 3 3B} & \textbf{Mean of five (P / I)} \\
\midrule
\rowcolor{bandcol} \multicolumn{7}{@{}l}{\textbf{\textit{Wrong zero-shot: \% put right}}} \\
Both texts & 274; 57.3 / 67.2 & 94; 66.0 / 69.1 & 231; 51.1 / 59.7 & 130; 59.2 / 62.3 & 199; 55.8 / 64.8 & 57.9 / 64.6 \\
Instructions only & 160; 28.1 / 54.4 & 80; 28.7 / 68.8 & 150; 14.7 / 54.0 & 104; 19.2 / 65.4 & 134; 23.9 / 55.2 & 22.9 / 59.6 \\
Passages only & 57; 45.6 / 40.4 & 32; 50.0 / 21.9 & 53; 43.4 / 32.1 & 40; 45.0 / 22.5 & 43; 37.2 / 34.9 & 44.2 / 30.3 \\
Neither & 123; 23.6 / 22.8 & 65; 32.3 / 24.6 & 111; 17.1 / 20.7 & 93; 17.2 / 23.7 & 102; 20.6 / 17.6 & 22.2 / 21.9 \\
\addlinespace[2pt]
\rowcolor{bandcol} \multicolumn{7}{@{}l}{\textbf{\textit{Right zero-shot: \% put wrong}}} \\
Both texts & 438; 17.8 / 12.1 & 618; 5.5 / 4.0 & 481; 9.6 / 8.7 & 582; 5.0 / 3.8 & 513; 9.4 / 7.6 & 9.5 / 7.2 \\
Instructions only & 110; 50.9 / 20.0 & 190; 17.9 / 5.8 & 120; 32.5 / 15.0 & 166; 22.3 / 6.0 & 136; 27.9 / 13.2 & 30.3 / 12.0 \\
Passages only & 52; 13.5 / 26.9 & 77; 10.4 / 10.4 & 56; 14.3 / 26.8 & 69; 5.8 / 18.8 & 66; 4.5 / 22.7 & 9.7 / 21.1 \\
Neither & 59; 50.8 / 52.5 & 117; 33.3 / 27.4 & 71; 39.4 / 40.8 & 89; 33.7 / 21.3 & 80; 37.5 / 36.2 & 39.0 / 35.7 \\
\bottomrule
\end{tabular}

\caption{Questions grouped by which text holds the fact, split by the student's greedy zero-shot answer. Each cell is the group's size, then the percent the five passages (P) put right / the five instructions (I) put right; for questions right zero-shot, the percent each put wrong.}
\label{tab:told-vs-found}
\end{table}

\subsection{Decontamination and the Question-Withheld Check} \label{app:decontam}

Table~\ref{tab:decontam} runs the 13-gram and 8-gram probes of Section~\ref{sec:corpus}, every instruction as the student sees it against every test question (stem and options). Table~\ref{tab:eightgrams} lists every phrase a test question shares with an instruction; each is a stock phrase of the domain. Table~\ref{tab:withheld} is the question-withheld check per student.

\begin{table}[htbp]
\centering\small
\begin{tabular}{@{}l r r r r@{}}
\toprule
\textbf{Task} & \textbf{$n$-gram} & \textbf{Test questions} & \textbf{Instructions} & \textbf{Max shared} \\
\midrule
MedQA & 13 & 0 of 1,273 & 0 & 0 \\
MedQA & 8 & 3 of 1,273 & 9 & 3 \\
MMLU Law & 13 & 0 of 540 & 0 & 0 \\
MMLU Law & 8 & 2 of 540 & 2 & 2 \\
MathQA & 13 & 0 of 2,975 & 0 & 0 \\
MathQA & 8 & 13 of 2,975 & 7 & 3 \\
\bottomrule
\end{tabular}
\caption{Shared 13-grams and 8-grams between the instructions and the test questions: the test questions flagged, the instructions flagged, and the largest number of 8-grams one test question shares with one instruction.}
\label{tab:decontam}
\end{table}

\begin{table}[htbp]
\centering\small
\begin{tabular}{@{}l p{3.4in} r r@{}}
\toprule
\textbf{Task} & \textbf{Shared phrase} & \textbf{Questions} & \textbf{Instructions} \\
\midrule
\rowcolor{bandcol} \multicolumn{4}{@{}l}{\textbf{\textit{MedQA}}} \\
& dopaminergic neurons in the substantia nigra pars compacta & 1 & 2 \\
& a non lactose fermenting oxidase positive gram negative & 1 & 1 \\
& non lactose fermenting oxidase positive gram negative rod & 1 & 1 \\
& holosystolic murmur at the left lower sternal border & 1 & 5 \\
\addlinespace[2pt]
\rowcolor{bandcol} \multicolumn{4}{@{}l}{\textbf{\textit{MMLU Law}}} \\
& law is not rationally related to any legitimate & 1 & 1 \\
& fee simple determinable with a possibility of reverter & 1 & 1 \\
\addlinespace[2pt]
\rowcolor{bandcol} \multicolumn{4}{@{}l}{\textbf{\textit{MathQA}}} \\
& integer less than or equal to x for & 2 & 2 \\
& surface area of a sphere is 4 r & 2 & 1 \\
& number of pens and same number of pencils & 4 & 1 \\
& sum of their products taken two at a & 1 & 1 \\
& of that is added to the weight the & 2 & 1 \\
& is equal to the side of the square & 2 & 1 \\
\bottomrule
\end{tabular}
\caption{Every phrase of eight or more words that a test question shares with an instruction, with the number of test questions and of instructions it appears in.}
\label{tab:eightgrams}
\end{table}

\begin{table}[htbp]
\centering\scriptsize
\setlength{\tabcolsep}{3pt}

\begin{tabular}{@{}l r l l r@{}}
\toprule
\textbf{Student} & \textbf{Options alone} & \textbf{Passages} & \textbf{Instructions} & \textbf{With the question} \\
\midrule
\rowcolor{bandcol} \multicolumn{5}{@{}l}{\textbf{\textit{MedQA}}} \\
Qwen 3.5 2B & 25.6 & 34.2 ($+$8.6 [$+$5.4, $+$11.9]) & 39.2 ($+$13.6 [$+$10.1, $+$16.9]) & 67.6 \\
Qwen 3.5 4B & 26.3 & 42.9 ($+$16.6 [$+$13.3, $+$19.7]) & 42.5 ($+$16.2 [$+$13.0, $+$19.2]) & 84.0 \\
Gemma 4 E2B & 19.6 & 40.9 ($+$21.3 [$+$18.3, $+$24.3]) & 37.6 ($+$18.0 [$+$14.8, $+$21.1]) & 69.4 \\
Gemma 4 E4B & 16.2 & 41.6 ($+$25.5 [$+$22.5, $+$28.4]) & 30.6 ($+$14.4 [$+$11.5, $+$17.0]) & 80.3 \\
Ministral 3 3B & 25.7 & 39.6 ($+$13.9 [$+$10.6, $+$17.2]) & 42.3 ($+$16.7 [$+$13.5, $+$20.0]) & 73.1 \\
\rowcolor{meanband} Mean of five & 22.7 & 39.8 & 38.4 & 74.9 \\
\addlinespace[2pt]
\rowcolor{bandcol} \multicolumn{5}{@{}l}{\textbf{\textit{MMLU Law}}} \\
Qwen 3.5 2B & 33.7 & -- & 37.6 ($+$3.9 [$-$1.3, $+$9.1]) & 40.9 \\
Qwen 3.5 4B & 39.3 & -- & 46.7 ($+$7.4 [$+$2.8, $+$11.9]) & 62.8 \\
Gemma 4 E2B & 32.0 & -- & 35.0 ($+$3.0 [$-$1.5, $+$7.4]) & 52.8 \\
Gemma 4 E4B & 33.1 & -- & 41.1 ($+$8.0 [$+$3.9, $+$12.4]) & 62.2 \\
Ministral 3 3B & 33.0 & -- & 38.5 ($+$5.6 [$+$0.7, $+$10.6]) & 52.4 \\
\rowcolor{meanband} Mean of five & 34.2 & -- & 39.8 & 54.2 \\
\addlinespace[2pt]
\rowcolor{bandcol} \multicolumn{5}{@{}l}{\textbf{\textit{MathQA}}} \\
Qwen 3.5 2B & 14.5 & -- & 13.4 ($-$1.0 [$-$2.7, $+$0.7]) & 79.8 \\
Qwen 3.5 4B & 12.7 & -- & 19.8 ($+$7.1 [$+$5.3, $+$8.9]) & 90.3 \\
Gemma 4 E2B & 10.9 & -- & 6.3 ($-$4.6 [$-$5.9, $-$3.3]) & 88.4 \\
Gemma 4 E4B & 6.8 & -- & 8.5 ($+$1.7 [$+$0.5, $+$3.0]) & 90.1 \\
Ministral 3 3B & 2.6 & -- & 3.4 ($+$0.8 [$-$0.1, $+$1.5]) & 84.1 \\
\rowcolor{meanband} Mean of five & 9.5 & -- & 10.3 & 86.5 \\
\bottomrule
\end{tabular}

\caption{The question-withheld check, greedy on test: percent of questions on which the student chooses the keyed option from the options alone, from the five retrieved passages (MedQA only), and from the five retrieved instructions, under the same ask, with the question for scale. A reply naming no option counts as not choosing the key. In parentheses, the paired difference from the options alone with its 95\% interval.}
\label{tab:withheld}
\end{table}

\FloatBarrier

\end{document}

%% file: math_commands.tex
\usepackage{amsmath,amsfonts,bm}

\def\eqref#1{equation~\ref{#1}}

\def\1{\bm{1}}

\DeclareMathAlphabet{\mathsfit}{\encodingdefault}{\sfdefault}{m}{sl}
\SetMathAlphabet{\mathsfit}{bold}{\encodingdefault}{\sfdefault}{bx}{n}

%% file: figures/figure1.tex
\begingroup
\hyphenpenalty=10000 \exhyphenpenalty=10000
\definecolor{figink}{HTML}{1A1A1A}
\definecolor{figgray}{HTML}{8C8C8C}      %
\definecolor{figline}{HTML}{9A9A9A}      %
\definecolor{figaccent}{HTML}{009E73}    %
\definecolor{fighl}{HTML}{CDEBDF}        %
\newlength{\qw}\setlength{\qw}{1.00in}   %
\newlength{\pw}\setlength{\pw}{1.52in}   %
\newlength{\lw}\setlength{\lw}{0.62in}   %
\newlength{\iw}\setlength{\iw}{1.20in}   %
\newlength{\colgap}\setlength{\colgap}{0.10in}

\tikzset{
  every node/.style={font=\scriptsize, text=figink, inner sep=0pt, outer sep=0pt, align=left},
  para/.style={text width=#1, inner xsep=3pt, inner ysep=1.5pt},
  hl/.style={fill=fighl, rounded corners=1pt},
  ptitle/.style={font=\footnotesize\bfseries},
  card/.style={draw=figline, line width=0.5pt, rounded corners=2pt, fill=white},
  ourcard/.style={draw=figaccent, line width=0.9pt, rounded corners=2pt, fill=white},
}
\begin{tikzpicture}

\node[ptitle, anchor=base west] (qt) at (0,0) {Question};
\node[para=\qw, anchor=north west] (q1) at ([yshift=-7pt]qt.base west) {A woman with ulcerative colitis \ldots\ bloody diarrhea \ldots\ fever, hypotension, and a distended, rigid abdomen \ldots\ Which is the most appropriate next step?};
\node[para=\qw, anchor=north west] (q2) at ([yshift=-3pt]q1.south west) {A.\ Double-contrast barium enema\\ B.\ Abdominal CT scan with contrast\\ C.\ CT angiography\\ D.\ Abdominal x-ray\\ E.\ Colonoscopy};
\begin{scope}[on background layer]
\node[card, fit=(q1)(q2), inner sep=4pt] (qcard) {};
\end{scope}

\coordinate (pcol) at ($(qcard.north east)+(\colgap,0)$);
\coordinate (icol) at ($(pcol)+(\pw+8pt+6pt+\colgap,0)$);
\node[ptitle, anchor=base west] (it) at (icol |- qt.base west) {Retrieved instructions};
\node[font=\scriptsize\bfseries, anchor=north west, inner xsep=3pt] (i0) at ($(icol |- pcol)+(4pt,-4pt)$) {Toxic megacolon in ulcerative colitis};
\node[para=\lw, font=\scriptsize\bfseries, anchor=north west] (l1) at ([yshift=-6pt]i0.south west) {Background knowledge};
\node[para=\iw, hl, anchor=north west] (t1) at ($(l1.north west)+(\lw,0)$) {Toxic megacolon is systemic toxicity plus colonic dilation ($>$6~cm), best seen on plain abdominal x-ray.};
\node[para=\iw, anchor=north west] (t1b) at ([yshift=-1pt]t1.south west) {Colonoscopy and contrast enemas are contraindicated \ldots};
\node[para=\lw, font=\scriptsize\bfseries, anchor=north west] (l2) at ([yshift=-6pt]t1b.south west -| l1.north west) {Procedure};
\node[para=\iw, hl, anchor=north west] (t2) at ($(l2.north west)+(\lw,0)$) {Obtain a plain abdominal x-ray immediately to assess colonic diameter and look for free air.};
\node[para=\iw, anchor=north west] (t2b) at ([yshift=-1pt]t2.south west) {Escalate to CT only if the diagnosis remains unclear \ldots};
\node[para=\lw, font=\scriptsize\bfseries, anchor=north west] (l3) at ([yshift=-6pt]t2b.south west -| l1.north west) {Common mistakes};
\node[para=\iw, hl, anchor=north west] (t3) at ($(l3.north west)+(\lw,0)$) {CT is not the initial test when megacolon is suspected and time and safety matter.};
\node[para=\iw, anchor=north west] (t3b) at ([yshift=-1pt]t3.south west) {Avoid colonoscopy and contrast enemas in suspected megacolon \ldots};
\begin{scope}[on background layer]
\node[fit=(i0)(l1)(t1)(t3b), inner sep=4pt, anchor=north west] (icard) at (icol |- pcol) {};
\node[ourcard, line width=0.5pt, fit=(icard), inner sep=0pt, shift={(6pt,-6pt)}] {};
\node[ourcard, line width=0.5pt, fit=(icard), inner sep=0pt, shift={(3pt,-3pt)}] {};
\node[ourcard, fit=(icard), inner sep=0pt] {};
\end{scope}

\node[ptitle, anchor=base west] (pt) at (pcol |- qt.base west) {Retrieved passages};
\coordinate (pse) at ($(pcol)+(\pw+8pt,0)$);
\begin{scope}[on background layer]
\node[fit=(pcol |- icard.north)(pse |- icard.south), inner sep=0pt] (pcard) {};
\node[card, fit=(pcard), inner sep=0pt, shift={(6pt,-6pt)}] {};
\node[card, fit=(pcard), inner sep=0pt, shift={(3pt,-3pt)}] {};
\node[card, fit=(pcard), inner sep=0pt] {};
\end{scope}
\begin{scope}
\clip ($(pcard.north west)+(1pt,-1pt)$) rectangle ($(pcard.south east)+(-1pt,3pt)$);
\node[para=\pw, font=\tiny, text=figgray, anchor=north west] (p1) at ($(pcol)+(4pt,-4pt)$) {\ldots\ The diagnosis of ulcerative colitis is almost always made endoscopically. Because the rectum is invariably involved, proctoscopy may be adequate to establish the diagnosis \ldots\ The earliest manifestation is mucosal edema, which appears as loss of the normal vascular pattern. In more severe disease, the mucosa is granular and friable \ldots\ Massive hemorrhage occurs with severe attacks of disease in 1\% of patients, and treatment for the disease usually stops the bleeding. However, if a patient requires 6--8 units of blood within 24--48~h, colectomy is indicated.};
\node[para=\pw, font=\tiny, hl, anchor=north west] (p2) at ([yshift=-1pt]p1.south west) {Toxic megacolon is defined as a transverse or right colon with a diameter of $>$6~cm.};
\node[para=\pw, font=\tiny, text=figgray, anchor=north west] (p3) at ([yshift=-1pt]p2.south west) {It occurs in about 5\% of attacks and can be triggered by electrolyte abnormalities and narcotics. About 50\% of acute dilations will resolve with medical therapy alone \ldots\ mucosal biopsy is often diagnostic in the chronic phase, biopsy in the acute phase will often reveal only nonspecific inflammation \ldots\ water-soluble contrast enema, or CT may be used to assess for an obstructing lesion and differentiate obstruction from pseudoobstruction. One of these diagnostic studies should be strongly considered if the patient does not have clear risk factors for pseudoobstruction \ldots\ Barium enema has been used to diagnose chronic ulcerative colitis and to determine the extent of disease. However, this modality is less sensitive than colonoscopy \ldots\ video capsule endoscopy; computed tomography (CT) scanning \ldots};
\end{scope}

\end{tikzpicture}
\endgroup

%% file: paper.bbl
\begin{thebibliography}{58}
\providecommand{\natexlab}[1]{#1}
\providecommand{\url}[1]{\texttt{#1}}
\expandafter\ifx\csname urlstyle\endcsname\relax
  \providecommand{\doi}[1]{doi: #1}\else
  \providecommand{\doi}{doi: \begingroup \urlstyle{rm}\Url}\fi

\bibitem[Ahmed et~al.(2026)Ahmed, Khan, Ahmad, Di, Liu, and
  Anwar]{ahmed2025retrievalofthought}
Ammar Ahmed, Azal~Ahmad Khan, Ayaan Ahmad, Sheng Di, Zirui Liu, and Ali Anwar.
\newblock {Retrieval-of-Thought}: Efficient reasoning via reusing thoughts.
\newblock In \emph{International Conference on Learning Representations}, 2026.
\newblock URL \url{https://arxiv.org/abs/2509.21743}.

\bibitem[Amini et~al.(2019)Amini, Gabriel, Lin, Koncel-Kedziorski, Choi, and
  Hajishirzi]{amini2019mathqainterpretablemathword}
Aida Amini, Saadia Gabriel, Shanchuan Lin, Rik Koncel-Kedziorski, Yejin Choi,
  and Hannaneh Hajishirzi.
\newblock {MathQA}: Towards interpretable math word problem solving with
  operation-based formalisms.
\newblock In \emph{Proceedings of the 2019 Conference of the North American
  Chapter of the Association for Computational Linguistics: Human Language
  Technologies, Volume 1 (Long and Short Papers)}, pp.\  2357--2367.
  Association for Computational Linguistics, 2019.
\newblock \doi{10.18653/v1/N19-1245}.
\newblock URL \url{https://aclanthology.org/N19-1245/}.

\bibitem[Arabzadeh et~al.(2026)Arabzadeh, Ma, Min, and
  Zaharia]{arabzadeh2026ragthinking}
Negar Arabzadeh, Wenjie Ma, Sewon Min, and Matei Zaharia.
\newblock {RAG} over thinking traces can improve reasoning tasks, 2026.
\newblock URL \url{https://arxiv.org/abs/2605.03344}.

\bibitem[Asai et~al.(2024)Asai, Wu, Wang, Sil, and
  Hajishirzi]{asai2024iclr-selfrag}
Akari Asai, Zeqiu Wu, Yizhong Wang, Avirup Sil, and Hannaneh Hajishirzi.
\newblock {Self-RAG}: Learning to retrieve, generate, and critique through
  self-reflection.
\newblock In \emph{International Conference on Learning Representations}, 2024.
\newblock URL \url{https://openreview.net/forum?id=hSyW5go0v8}.

\bibitem[Belcak et~al.(2025)Belcak, Heinrich, Diao, Fu, Dong, Muralidharan,
  Lin, and Molchanov]{belcak2025smalllanguagemodelsfuture}
Peter Belcak, Greg Heinrich, Shizhe Diao, Yonggan Fu, Xin Dong, Saurav
  Muralidharan, Yingyan~Celine Lin, and Pavlo Molchanov.
\newblock Small language models are the future of agentic {AI}, 2025.
\newblock URL \url{https://arxiv.org/abs/2506.02153}.

\bibitem[Brown et~al.(2024)Brown, Juravsky, Ehrlich, Clark, Le, R{\'e}, and
  Mirhoseini]{brown2024monkeys}
Bradley Brown, Jordan Juravsky, Ryan Ehrlich, Ronald Clark, Quoc~V. Le,
  Christopher R{\'e}, and Azalia Mirhoseini.
\newblock Large language monkeys: Scaling inference compute with repeated
  sampling, 2024.
\newblock URL \url{https://arxiv.org/abs/2407.21787}.

\bibitem[Brown et~al.(2020)Brown, Mann, Ryder, Subbiah, Kaplan, Dhariwal,
  Neelakantan, Shyam, Sastry, Askell, Agarwal, Herbert-Voss, Krueger, Henighan,
  Child, Ramesh, Ziegler, Wu, Winter, Hesse, Chen, Sigler, Litwin, Gray, Chess,
  Clark, Berner, McCandlish, Radford, Sutskever, and
  Amodei]{brown2020languagemodelsfewshotlearners}
Tom~B. Brown, Benjamin Mann, Nick Ryder, Melanie Subbiah, Jared Kaplan,
  Prafulla Dhariwal, Arvind Neelakantan, Pranav Shyam, Girish Sastry, Amanda
  Askell, Sandhini Agarwal, Ariel Herbert-Voss, Gretchen Krueger, Tom Henighan,
  Rewon Child, Aditya Ramesh, Daniel~M. Ziegler, Jeffrey Wu, Clemens Winter,
  Christopher Hesse, Mark Chen, Eric Sigler, Mateusz Litwin, Scott Gray,
  Benjamin Chess, Jack Clark, Christopher Berner, Sam McCandlish, Alec Radford,
  Ilya Sutskever, and Dario Amodei.
\newblock Language models are few-shot learners.
\newblock In \emph{Advances in Neural Information Processing Systems},
  volume~33, 2020.
\newblock URL \url{https://arxiv.org/abs/2005.14165}.

\bibitem[Chen et~al.(2024)Chen, Zaharia, and Zou]{chen2024tmlr-frugalgpt}
Lingjiao Chen, Matei Zaharia, and James Zou.
\newblock {FrugalGPT}: How to use large language models while reducing cost and
  improving performance.
\newblock \emph{Transactions on Machine Learning Research}, 2024.
\newblock URL \url{https://openreview.net/forum?id=cSimKw5p6R}.

\bibitem[Chen et~al.(2021)Chen, Tworek, Jun, Yuan, de~Oliveira~Pinto, Kaplan,
  Edwards, Burda, Joseph, Brockman, et~al.]{chen2021codex}
Mark Chen, Jerry Tworek, Heewoo Jun, Qiming Yuan, Henrique~Ponde
  de~Oliveira~Pinto, Jared Kaplan, Harri Edwards, Yuri Burda, Nicholas Joseph,
  Greg Brockman, et~al.
\newblock Evaluating large language models trained on code, 2021.
\newblock URL \url{https://arxiv.org/abs/2107.03374}.

\bibitem[Chowdhery et~al.(2023)Chowdhery, Narang, Devlin, Bosma, Mishra,
  et~al.]{chowdhery2023palm}
Aakanksha Chowdhery, Sharan Narang, Jacob Devlin, Maarten Bosma, Gaurav Mishra,
  et~al.
\newblock {PaLM}: Scaling language modeling with pathways.
\newblock \emph{Journal of Machine Learning Research}, 24\penalty0
  (240):\penalty0 1--113, 2023.
\newblock URL \url{http://jmlr.org/papers/v24/22-1144.html}.

\bibitem[Cobbe et~al.(2021)Cobbe, Kosaraju, Bavarian, Chen, Jun, Kaiser,
  Plappert, Tworek, Hilton, Nakano, Hesse, and
  Schulman]{cobbe2021trainingverifierssolvemath}
Karl Cobbe, Vineet Kosaraju, Mohammad Bavarian, Mark Chen, Heewoo Jun, Lukasz
  Kaiser, Matthias Plappert, Jerry Tworek, Jacob Hilton, Reiichiro Nakano,
  Christopher Hesse, and John Schulman.
\newblock Training verifiers to solve math word problems, 2021.
\newblock URL \url{https://arxiv.org/abs/2110.14168}.

\bibitem[Frantar et~al.(2023)Frantar, Ashkboos, Hoefler, and
  Alistarh]{frantar2023gptqaccurateposttrainingquantization}
Elias Frantar, Saleh Ashkboos, Torsten Hoefler, and Dan Alistarh.
\newblock {OPTQ}: Accurate post-training quantization for generative
  pre-trained transformers.
\newblock In \emph{International Conference on Learning Representations}, 2023.
\newblock URL \url{https://openreview.net/forum?id=tcbBPnfwxS}.

\bibitem[Fu et~al.(2023)Fu, Peng, Ou, Sabharwal, and
  Khot]{fu2023specializingsmallerlanguagemodels}
Yao Fu, Hao Peng, Litu Ou, Ashish Sabharwal, and Tushar Khot.
\newblock Specializing smaller language models towards multi-step reasoning.
\newblock In \emph{Proceedings of the 40th International Conference on Machine
  Learning}, volume 202 of \emph{Proceedings of Machine Learning Research},
  pp.\  10421--10430. PMLR, 2023.
\newblock URL \url{https://proceedings.mlr.press/v202/fu23d.html}.

\bibitem[{Gemma Team}(2025)]{gemma3_2025}
{Gemma Team}.
\newblock Gemma 3 technical report.
\newblock \emph{arXiv preprint arXiv:2503.19786}, 2025.

\bibitem[{Gemma Team}(2026)]{gemmateam2026gemma4}
{Gemma Team}.
\newblock {Gemma 4} technical report, 2026.
\newblock URL \url{https://arxiv.org/abs/2607.02770}.

\bibitem[Grattafiori et~al.(2024)]{grattafiori2024llama3herd}
Aaron Grattafiori et~al.
\newblock The {Llama} 3 herd of models, 2024.
\newblock URL \url{https://arxiv.org/abs/2407.21783}.

\bibitem[Guo et~al.(2025)Guo, Yang, Zhang, Song, Wang,
  et~al.]{deepseekai2025deepseekr1incentivizingreasoningcapability}
Daya Guo, Dejian Yang, Haowei Zhang, Junxiao Song, Peiyi Wang, et~al.
\newblock {DeepSeek-R1} incentivizes reasoning in {LLMs} through reinforcement
  learning.
\newblock \emph{Nature}, 645:\penalty0 633--638, 2025.
\newblock \doi{10.1038/s41586-025-09422-z}.

\bibitem[Hendrycks et~al.(2021{\natexlab{a}})Hendrycks, Burns, Basart, Zou,
  Mazeika, Song, and
  Steinhardt]{hendrycks2021measuringmassivemultitasklanguage}
Dan Hendrycks, Collin Burns, Steven Basart, Andy Zou, Mantas Mazeika, Dawn
  Song, and Jacob Steinhardt.
\newblock Measuring massive multitask language understanding.
\newblock In \emph{International Conference on Learning Representations},
  2021{\natexlab{a}}.
\newblock URL \url{https://arxiv.org/abs/2009.03300}.

\bibitem[Hendrycks et~al.(2021{\natexlab{b}})Hendrycks, Burns, Kadavath, Arora,
  Basart, Tang, Song, and
  Steinhardt]{hendrycks2021measuringmathematicalproblemsolving}
Dan Hendrycks, Collin Burns, Saurav Kadavath, Akul Arora, Steven Basart, Eric
  Tang, Dawn Song, and Jacob Steinhardt.
\newblock Measuring mathematical problem solving with the {MATH} dataset.
\newblock In \emph{Proceedings of the Neural Information Processing Systems
  Track on Datasets and Benchmarks}, 2021{\natexlab{b}}.
\newblock URL \url{https://arxiv.org/abs/2103.03874}.

\bibitem[Ho et~al.(2023)Ho, Schmid, and
  Yun]{ho2023largelanguagemodelsreasoning}
Namgyu Ho, Laura Schmid, and Se-Young Yun.
\newblock Large language models are reasoning teachers.
\newblock In \emph{Proceedings of the 61st Annual Meeting of the Association
  for Computational Linguistics (Volume 1: Long Papers)}, pp.\  14852--14882,
  2023.
\newblock \doi{10.18653/v1/2023.acl-long.830}.
\newblock URL \url{https://aclanthology.org/2023.acl-long.830/}.

\bibitem[Hsieh et~al.(2023)Hsieh, Li, kuan Yeh, Nakhost, Fujii, Ratner,
  Krishna, Lee, and Pfister]{hsieh2023distillingstepbystepoutperforminglarger}
Cheng-Yu Hsieh, Chun-Liang Li, Chih kuan Yeh, Hootan Nakhost, Yasuhisa Fujii,
  Alex Ratner, Ranjay Krishna, Chen-Yu Lee, and Tomas Pfister.
\newblock Distilling step-by-step! outperforming larger language models with
  less training data and smaller model sizes.
\newblock In \emph{Findings of the Association for Computational Linguistics:
  ACL 2023}, pp.\  8003--8017, 2023.
\newblock \doi{10.18653/v1/2023.findings-acl.507}.
\newblock URL \url{https://aclanthology.org/2023.findings-acl.507/}.

\bibitem[Jeong et~al.(2026)Jeong, Jung, Hwang, Kim, and
  Kang]{jeong2025thoughttemplates}
Soyeong Jeong, Taehee Jung, Sung~Ju Hwang, Joo-Kyung Kim, and Dongyeop Kang.
\newblock When thoughts meet facts: Reusable reasoning for long-context {LM}s.
\newblock In \emph{Findings of the {A}ssociation for {C}omputational
  {L}inguistics: {ACL} 2026}, pp.\  1625--1646, San Diego, California, United
  States, July 2026. Association for Computational Linguistics.
\newblock \doi{10.18653/v1/2026.findings-acl.81}.
\newblock URL \url{https://aclanthology.org/2026.findings-acl.81/}.

\bibitem[Jin et~al.(2021)Jin, Pan, Oufattole, Weng, Fang, and
  Szolovits]{jin2020diseasedoespatienthave}
Di~Jin, Eileen Pan, Nassim Oufattole, Wei-Hung Weng, Hanyi Fang, and Peter
  Szolovits.
\newblock What disease does this patient have? {A} large-scale open domain
  question answering dataset from medical exams.
\newblock \emph{Applied Sciences}, 11\penalty0 (14):\penalty0 6421, 2021.
\newblock \doi{10.3390/app11146421}.
\newblock URL \url{https://www.mdpi.com/2076-3417/11/14/6421}.

\bibitem[Joren et~al.(2025)Joren, Zhang, Ferng, Juan, Taly, and
  Rashtchian]{joren2025iclr-sufficient}
Hailey Joren, Jianyi Zhang, Chun-Sung Ferng, Da-Cheng Juan, Ankur Taly, and
  Cyrus Rashtchian.
\newblock Sufficient context: A new lens on retrieval augmented generation
  systems.
\newblock In \emph{International Conference on Learning Representations}, 2025.
\newblock URL \url{https://openreview.net/forum?id=Jjr2Odj8DJ}.

\bibitem[Kang et~al.(2023)Kang, Lee, Baek, Kawaguchi, and
  Hwang]{kang2023knowledgeaugmentedreasoningdistillationsmall}
Minki Kang, Seanie Lee, Jinheon Baek, Kenji Kawaguchi, and Sung~Ju Hwang.
\newblock Knowledge-augmented reasoning distillation for small language models
  in knowledge-intensive tasks.
\newblock In \emph{Advances in Neural Information Processing Systems},
  volume~36, 2023.
\newblock URL \url{https://arxiv.org/abs/2305.18395}.

\bibitem[Karpukhin et~al.(2020)Karpukhin, Oguz, Min, Lewis, Wu, Edunov, Chen,
  and Yih]{karpukhin-etal-2020-dense}
Vladimir Karpukhin, Barlas Oguz, Sewon Min, Patrick Lewis, Ledell Wu, Sergey
  Edunov, Danqi Chen, and Wen-tau Yih.
\newblock Dense passage retrieval for open-domain question answering.
\newblock In \emph{Proceedings of the 2020 Conference on Empirical Methods in
  Natural Language Processing (EMNLP)}, pp.\  6769--6781, Online, November
  2020. Association for Computational Linguistics.
\newblock \doi{10.18653/v1/2020.emnlp-main.550}.
\newblock URL \url{https://aclanthology.org/2020.emnlp-main.550/}.

\bibitem[Katz et~al.(2024)Katz, Bommarito, Gao, and Arredondo]{katz2024bar}
Daniel~Martin Katz, Michael~James Bommarito, Shang Gao, and Pablo Arredondo.
\newblock {GPT-4} passes the bar exam.
\newblock \emph{Philosophical Transactions of the Royal Society A:
  Mathematical, Physical and Engineering Sciences}, 382\penalty0 (2270), 2024.
\newblock \doi{10.1098/rsta.2023.0254}.

\bibitem[Kotha et~al.(2024)Kotha, Springer, and
  Raghunathan]{kotha2024iclr-understanding}
Suhas Kotha, Jacob~Mitchell Springer, and Aditi Raghunathan.
\newblock Understanding catastrophic forgetting in language models via implicit
  inference.
\newblock In \emph{International Conference on Learning Representations}, 2024.

\bibitem[Kwon et~al.(2024)Kwon, iunn Ong, Kang, Moon, Lee, Hwang, Sohn, Sim,
  Lee, and Yeo]{kwon2024largelanguagemodelsclinical}
Taeyoon Kwon, Kai~Tzu iunn Ong, Dongjin Kang, Seungjun Moon, Jeong~Ryong Lee,
  Dosik Hwang, Beomseok Sohn, Yongsik Sim, Dongha Lee, and Jinyoung Yeo.
\newblock Large language models are clinical reasoners: Reasoning-aware
  diagnosis framework with prompt-generated rationales.
\newblock In \emph{Proceedings of the AAAI Conference on Artificial
  Intelligence}, volume~38, pp.\  18417--18425, 2024.
\newblock \doi{10.1609/aaai.v38i16.29802}.
\newblock URL \url{https://doi.org/10.1609/aaai.v38i16.29802}.

\bibitem[Lewis et~al.(2020)Lewis, Perez, Piktus, Petroni, Karpukhin, Goyal,
  Küttler, Lewis, tau Yih, Rocktäschel, Riedel, and
  Kiela]{lewis2021retrievalaugmentedgenerationknowledgeintensivenlp}
Patrick Lewis, Ethan Perez, Aleksandra Piktus, Fabio Petroni, Vladimir
  Karpukhin, Naman Goyal, Heinrich Küttler, Mike Lewis, Wen tau Yih, Tim
  Rocktäschel, Sebastian Riedel, and Douwe Kiela.
\newblock Retrieval-augmented generation for knowledge-intensive nlp tasks.
\newblock In \emph{Advances in Neural Information Processing Systems},
  volume~33, 2020.
\newblock URL \url{https://arxiv.org/abs/2005.11401}.

\bibitem[Li et~al.(2022)Li, Chen, Shen, Chen, Zhang, Li, Wang, Qian, Peng, Mao,
  Chen, and Yan]{li2022explanationslargelanguagemodels}
Shiyang Li, Jianshu Chen, Yelong Shen, Zhiyu Chen, Xinlu Zhang, Zekun Li, Hong
  Wang, Jing Qian, Baolin Peng, Yi~Mao, Wenhu Chen, and Xifeng Yan.
\newblock Explanations from large language models make small reasoners better,
  2022.
\newblock URL \url{https://arxiv.org/abs/2210.06726}.

\bibitem[Li et~al.(2026)Li, Wu, and Boulet]{li2026trainingfree}
Yuran Li, Di~Wu, and Benoit Boulet.
\newblock A training-free regeneration paradigm: Contrastive reflection memory
  guided self-verification and self-improvement, 2026.

\bibitem[Liu et~al.(2022)Liu, Liu, Lu, Welleck, West, Bras, Choi, and
  Hajishirzi]{liu2022generatedknowledge}
Jiacheng Liu, Alisa Liu, Ximing Lu, Sean Welleck, Peter West, Ronan~Le Bras,
  Yejin Choi, and Hannaneh Hajishirzi.
\newblock Generated knowledge prompting for commonsense reasoning.
\newblock In \emph{Proceedings of the 60th Annual Meeting of the Association
  for Computational Linguistics (Volume 1: Long Papers)}, pp.\  3154--3169,
  2022.
\newblock \doi{10.18653/v1/2022.acl-long.225}.
\newblock URL \url{https://aclanthology.org/2022.acl-long.225/}.

\bibitem[Lu et~al.(2024)Lu, Li, Cheng, Ding, Huang, and
  Qiu]{lu-etal-2024-scaling}
Xingyu Lu, Xiaonan Li, Qinyuan Cheng, Kai Ding, Xuanjing Huang, and Xipeng Qiu.
\newblock Scaling laws for fact memorization of large language models.
\newblock In Yaser Al-Onaizan, Mohit Bansal, and Yun-Nung Chen (eds.),
  \emph{Findings of the Association for Computational Linguistics: EMNLP 2024},
  pp.\  11263--11282, Miami, Florida, USA, November 2024. Association for
  Computational Linguistics.
\newblock \doi{10.18653/v1/2024.findings-emnlp.658}.
\newblock URL \url{https://aclanthology.org/2024.findings-emnlp.658/}.

\bibitem[Magister et~al.(2023)Magister, Mallinson, Adamek, Malmi, and
  Severyn]{magister2023teachingsmalllanguagemodels}
Lucie~Charlotte Magister, Jonathan Mallinson, Jakub Adamek, Eric Malmi, and
  Aliaksei Severyn.
\newblock Teaching small language models to reason.
\newblock In \emph{Proceedings of the 61st Annual Meeting of the Association
  for Computational Linguistics (Volume 2: Short Papers)}, pp.\  1773--1781,
  2023.
\newblock \doi{10.18653/v1/2023.acl-short.151}.
\newblock URL \url{https://aclanthology.org/2023.acl-short.151/}.

\bibitem[Mallen et~al.(2023)Mallen, Asai, Zhong, Das, Khashabi, and
  Hajishirzi]{mallen2023trust}
Alex Mallen, Akari Asai, Victor Zhong, Rajarshi Das, Daniel Khashabi, and
  Hannaneh Hajishirzi.
\newblock When not to trust language models: Investigating effectiveness of
  parametric and non-parametric memories.
\newblock In \emph{Proceedings of the 61st Annual Meeting of the Association
  for Computational Linguistics (Volume 1: Long Papers)}, pp.\  9802--9822,
  2023.
\newblock \doi{10.18653/v1/2023.acl-long.546}.
\newblock URL \url{https://aclanthology.org/2023.acl-long.546/}.

\bibitem[{Mistral AI}(2026)]{mistral2026ministral3}
{Mistral AI}.
\newblock Ministral 3, 2026.

\bibitem[{OpenAI}(2025)]{openai2025gpt5}
{OpenAI}.
\newblock {OpenAI GPT-5} system card, 2025.
\newblock URL \url{https://arxiv.org/abs/2601.03267}.

\bibitem[Pham et~al.(2025)Pham, Nguyen, Yoon, Lai, Dernoncourt, and
  Bui]{pham2024slimlmefficientsmalllanguage}
Thang~M. Pham, Phat~T. Nguyen, Seunghyun Yoon, Viet~Dac Lai, Franck
  Dernoncourt, and Trung Bui.
\newblock {SlimLM}: An efficient small language model for on-device document
  assistance.
\newblock In \emph{Proceedings of the 63rd Annual Meeting of the Association
  for Computational Linguistics (Volume 3: System Demonstrations)}, pp.\
  436--447, 2025.
\newblock \doi{10.18653/v1/2025.acl-demo.42}.
\newblock URL \url{https://aclanthology.org/2025.acl-demo.42/}.

\bibitem[{Qwen Team}(2026)]{qwenteam2026qwen35}
{Qwen Team}.
\newblock {Qwen3.5}: Towards native multimodal agents.
\newblock Blog post, February 2026.
\newblock URL \url{https://qwen.ai/blog?id=qwen3.5}.

\bibitem[Ram et~al.(2023)Ram, Levine, Dalmedigos, Muhlgay, Shashua,
  Leyton-Brown, and Shoham]{ram-etal-2023-context}
Ori Ram, Yoav Levine, Itay Dalmedigos, Dor Muhlgay, Amnon Shashua, Kevin
  Leyton-Brown, and Yoav Shoham.
\newblock In-context retrieval-augmented language models.
\newblock \emph{Transactions of the Association for Computational Linguistics},
  11:\penalty0 1316--1331, 2023.
\newblock \doi{10.1162/tacl\_a\_00605}.
\newblock URL \url{https://aclanthology.org/2023.tacl-1.75/}.

\bibitem[Roberts et~al.(2020)Roberts, Raffel, and
  Shazeer]{roberts-etal-2020-much}
Adam Roberts, Colin Raffel, and Noam Shazeer.
\newblock How much knowledge can you pack into the parameters of a language
  model?
\newblock In \emph{Proceedings of the 2020 Conference on Empirical Methods in
  Natural Language Processing (EMNLP)}, pp.\  5418--5426. Association for
  Computational Linguistics, 2020.
\newblock \doi{10.18653/v1/2020.emnlp-main.437}.

\bibitem[Singhal et~al.(2025)Singhal, Tu, Gottweis, Sayres, Wulczyn, Amin, Hou,
  Clark, Pfohl, Cole-Lewis, Neal, Rashid, Schaekermann, Wang, Dash, Chen, Shah,
  Lachgar, Mansfield, Prakash, Green, Dominowska, Ag{\"u}era~y Arcas,
  Toma{\v{s}}ev, Liu, Wong, Semturs, Mahdavi, Barral, Webster, Corrado, Matias,
  Azizi, Karthikesalingam, and Natarajan]{singhal2025expert}
Karan Singhal, Tao Tu, Juraj Gottweis, Rory Sayres, Ellery Wulczyn, Mohamed
  Amin, Le~Hou, Kevin Clark, Stephen~R. Pfohl, Heather Cole-Lewis, Darlene
  Neal, Qazi~Mamunur Rashid, Mike Schaekermann, Amy Wang, Dev Dash, Jonathan~H.
  Chen, Nigam~H. Shah, Sami Lachgar, Philip~A. Mansfield, Sushant Prakash,
  Bradley Green, Ewa Dominowska, Blaise Ag{\"u}era~y Arcas, Nenad
  Toma{\v{s}}ev, Yun Liu, Renee Wong, Christopher Semturs, S.~Sara Mahdavi,
  Joelle~K. Barral, Dale~R. Webster, Greg~S. Corrado, Yossi Matias, Shekoofeh
  Azizi, Alan Karthikesalingam, and Vivek Natarajan.
\newblock Toward expert-level medical question answering with large language
  models.
\newblock \emph{Nature Medicine}, 31:\penalty0 943--950, 2025.
\newblock \doi{10.1038/s41591-024-03423-7}.

\bibitem[Su et~al.(2026)Su, Zhang, Luo, Liu, and Huang]{su2025mistake}
Xuanbo Su, Yingfang Zhang, Hao Luo, Xiaoteng Liu, and Leo Huang.
\newblock Mistake notebook learning: Batch-clustered failures for training-free
  agent adaptation.
\newblock In \emph{Findings of the Association for Computational Linguistics:
  ACL 2026}, pp.\  14629--14645. Association for Computational Linguistics,
  2026.
\newblock \doi{10.18653/v1/2026.findings-acl.719}.

\bibitem[Sun et~al.(2024)Sun, Jiang, Wang, Hou, Zhang, Xie, and
  Huang]{sun-etal-2024-retrieved}
Hao Sun, Yong Jiang, Bo~Wang, Yingyan Hou, Yan Zhang, Pengjun Xie, and Fei
  Huang.
\newblock Retrieved in-context principles from previous mistakes.
\newblock In \emph{Proceedings of the 2024 Conference on Empirical Methods in
  Natural Language Processing}, pp.\  8155--8169, Miami, Florida, USA, November
  2024. Association for Computational Linguistics.
\newblock \doi{10.18653/v1/2024.emnlp-main.465}.

\bibitem[Wang et~al.(2023)Wang, Wei, Schuurmans, Le, Chi, Narang, Chowdhery,
  and Zhou]{wang2023selfconsistencyimproveschainthought}
Xuezhi Wang, Jason Wei, Dale Schuurmans, Quoc Le, Ed~Chi, Sharan Narang,
  Aakanksha Chowdhery, and Denny Zhou.
\newblock Self-consistency improves chain of thought reasoning in language
  models.
\newblock In \emph{International Conference on Learning Representations}, 2023.
\newblock URL \url{https://arxiv.org/abs/2203.11171}.

\bibitem[Wang et~al.(2024)Wang, Ma, and
  Chen]{wang2025augmentingblackboxllmsmedical}
Yubo Wang, Xueguang Ma, and Wenhu Chen.
\newblock Augmenting black-box {LLMs} with medical textbooks for biomedical
  question answering.
\newblock In \emph{Findings of the Association for Computational Linguistics:
  EMNLP 2024}, pp.\  1754--1770, 2024.
\newblock \doi{10.18653/v1/2024.findings-emnlp.95}.
\newblock URL \url{https://aclanthology.org/2024.findings-emnlp.95/}.

\bibitem[Wei et~al.(2022)Wei, Wang, Schuurmans, Bosma, Ichter, Xia, Chi, Le,
  and Zhou]{wei2023chainofthoughtpromptingelicitsreasoning}
Jason Wei, Xuezhi Wang, Dale Schuurmans, Maarten Bosma, Brian Ichter, Fei Xia,
  Ed~Chi, Quoc Le, and Denny Zhou.
\newblock Chain-of-thought prompting elicits reasoning in large language
  models.
\newblock In \emph{Advances in Neural Information Processing Systems}, 2022.
\newblock URL \url{https://arxiv.org/abs/2201.11903}.

\bibitem[Xiao et~al.(2025)Xiao, Cai, Zhao, Lin, Zeng, Zhou, Zheng, Han, Liu,
  and Sun]{xiao2025densing}
Chaojun Xiao, Jie Cai, Weilin Zhao, Biyuan Lin, Guoyang Zeng, Jie Zhou, Zhi
  Zheng, Xu~Han, Zhiyuan Liu, and Maosong Sun.
\newblock Densing law of {LLMs}.
\newblock \emph{Nature Machine Intelligence}, 7\penalty0 (11):\penalty0
  1823--1833, 2025.
\newblock \doi{10.1038/s42256-025-01137-0}.

\bibitem[Xiao et~al.(2023)Xiao, Lin, Seznec, Wu, Demouth, and
  Han]{pmlr-v202-xiao23c}
Guangxuan Xiao, Ji~Lin, Mickael Seznec, Hao Wu, Julien Demouth, and Song Han.
\newblock {S}mooth{Q}uant: Accurate and efficient post-training quantization
  for large language models.
\newblock In \emph{Proceedings of the 40th International Conference on Machine
  Learning}, volume 202 of \emph{Proceedings of Machine Learning Research},
  pp.\  38087--38099. PMLR, 2023.
\newblock URL \url{https://proceedings.mlr.press/v202/xiao23c.html}.

\bibitem[Yang et~al.(2024)Yang, Yu, Zhang, Cao, Xu, Zhang, Gonzalez, and
  Cui]{yang2024bufferofthoughts}
Ling Yang, Zhaochen Yu, Tianjun Zhang, Shiyi Cao, Minkai Xu, Wentao Zhang,
  Joseph~E. Gonzalez, and Bin Cui.
\newblock {Buffer of Thoughts}: Thought-augmented reasoning with large language
  models.
\newblock In \emph{Advances in Neural Information Processing Systems}, 2024.
\newblock URL \url{https://arxiv.org/abs/2406.04271}.

\bibitem[Yang et~al.(2025{\natexlab{a}})Yang, Yu, Cui, and
  Wang]{yang2025reasonflux}
Ling Yang, Zhaochen Yu, Bin Cui, and Mengdi Wang.
\newblock {ReasonFlux}: Hierarchical {LLM} reasoning via scaling thought
  templates, 2025{\natexlab{a}}.
\newblock URL \url{https://arxiv.org/abs/2502.06772}.

\bibitem[Yang et~al.(2025{\natexlab{b}})Yang, Yu, Zhang, Xu, Gonzalez, Cui, and
  Yan]{yang2025supercorrect}
Ling Yang, Zhaochen Yu, Tianjun Zhang, Minkai Xu, Joseph~E. Gonzalez, Bin Cui,
  and Shuicheng Yan.
\newblock {SuperCorrect}: Advancing small {LLM} reasoning with thought template
  distillation and self-correction.
\newblock In \emph{International Conference on Learning Representations},
  2025{\natexlab{b}}.
\newblock URL \url{https://openreview.net/forum?id=PyjZO7oSw2}.

\bibitem[Yu et~al.(2023)Yu, Iter, Wang, Xu, Ju, Sanyal, Zhu, Zeng, and
  Jiang]{yu2023generateretrievelargelanguage}
Wenhao Yu, Dan Iter, Shuohang Wang, Yichong Xu, Mingxuan Ju, Soumya Sanyal,
  Chenguang Zhu, Michael Zeng, and Meng Jiang.
\newblock Generate rather than retrieve: Large language models are strong
  context generators.
\newblock In \emph{International Conference on Learning Representations}, 2023.
\newblock URL \url{https://arxiv.org/abs/2209.10063}.

\bibitem[Zhang et~al.(2024)Zhang, Patil, Jain, Shen, Zaharia, Stoica, and
  Gonzalez]{zhang2024raft}
Tianjun Zhang, Shishir~G. Patil, Naman Jain, Sheng Shen, Matei Zaharia, Ion
  Stoica, and Joseph~E. Gonzalez.
\newblock {RAFT}: Adapting language model to domain specific {RAG}.
\newblock In \emph{First Conference on Language Modeling (COLM)}, 2024.
\newblock URL \url{https://openreview.net/forum?id=rzQGHXNReU}.

\bibitem[Zhao et~al.(2024)Zhao, Zhou, and Zhu]{zhao-etal-2024-probe}
Yichun Zhao, Shuheng Zhou, and Huijia Zhu.
\newblock Probe then retrieve and reason: Distilling probing and reasoning
  capabilities into smaller language models.
\newblock In \emph{Proceedings of the 2024 Joint International Conference on
  Computational Linguistics, Language Resources and Evaluation (LREC-COLING
  2024)}, pp.\  13026--13032, Torino, Italia, May 2024. ELRA and ICCL.
\newblock URL \url{https://aclanthology.org/2024.lrec-main.1140/}.

\bibitem[Zhou et~al.(2024)Zhou, Pujara, Ren, Chen, Cheng, Le, Chi, Zhou,
  Mishra, and Zheng]{zhou2024selfdiscover}
Pei Zhou, Jay Pujara, Xiang Ren, Xinyun Chen, Heng-Tze Cheng, Quoc~V. Le, Ed~H.
  Chi, Denny Zhou, Swaroop Mishra, and Huaixiu~Steven Zheng.
\newblock {Self-Discover}: Large language models self-compose reasoning
  structures.
\newblock In \emph{Advances in Neural Information Processing Systems}, 2024.
\newblock URL \url{https://arxiv.org/abs/2402.03620}.

\bibitem[Zhu et~al.(2024)Zhu, Xue, Chen, Zhou, Tang, Schuurmans, and
  Dai]{zhu2024learnrules}
Zhaocheng Zhu, Yuan Xue, Xinyun Chen, Denny Zhou, Jian Tang, Dale Schuurmans,
  and Hanjun Dai.
\newblock Large language models can learn rules, 2024.
\newblock URL \url{https://arxiv.org/abs/2310.07064}.

\end{thebibliography}
